\documentclass[11pt, a4paper, logo, copyright]{googledeepmind}

\usepackage[authoryear, sort&compress, round]{natbib}

\usepackage{soul}

\usepackage{multicol}
\usepackage{multirow}
\usepackage{subcaption}

\usepackage{listings}

\usepackage{tcolorbox}

\definecolor{lightgray}{gray}{0.93}
\sethlcolor{lightgray}
\newcommand{\code}[1]{\hl{\texttt{#1}}}
\newcommand{\ag}{AlphaGeometry}  
\renewcommand{\cite}[1]{\citep{#1}}

\title{Gold-medalist Performance in Solving Olympiad Geometry with AlphaGeometry2}

\correspondingauthor{cyuri@google.com, thtrieu@google.com, thangluong@google.com\\
*, $\dagger$: Equal contributions}


\author[*,1,$\diamond$]{Yuri Chervonyi}
\author[*,1,$\diamond$]{Trieu H. Trinh}
\author[$\dagger$,1,2]{Miroslav Olšák}
\author[$\dagger$,1]{Xiaomeng Yang}
\author[1,3]{Hoang Nguyen}
\author{Marcelo Menegali}
\author[1,4]{Junehyuk Jung}
\author[1,5]{Junsu Kim}
\author{Vikas Verma}
\author[1]{Quoc V. Le}
\author[1,$\diamond$]{Thang Luong}

\affil[1]{Google DeepMind}
\affil[2]{University of Cambridge}
\affil[3]{Georgia Institute of Technology}
\affil[4]{Brown University}
\affil[5]{Seoul National University \protect\\ This work was conducted entirely at Google DeepMind by all authors.}

\begin{abstract}
We present AlphaGeometry2 (AG2), a significantly improved version of AlphaGeometry introduced in \cite{Trinh2024}, which has now surpassed an average gold medalist in solving Olympiad geometry problems. To achieve this, we first extend the original AlphaGeometry language to tackle problems involving movements of objects, and problems containing linear equations of angles, ratios, and distances. This, together with support for non-constructive problems, has markedly improved the coverage rate of the AlphaGeometry language on International Math Olympiads (IMO) 2000-2024 geometry problems from 66\% to 88\%. The search process of AG2 has also been greatly improved through the use of Gemini architecture for better language modeling, and a novel knowledge-sharing mechanism that enables effective communication between search trees. Together with further enhancements to the symbolic engine and synthetic data generation, we have significantly boosted the overall solving rate of AG to 84\% on \textit{all} geometry problems over the last 25 years, compared to 54\% previously. AG2 was also part of the system that achieved the silver-medal standard at IMO 2024 \url{https://deepmind.google/blog/ai-solves-imo-problems-at-silver-medal-level/}. Finally, we report progress towards using AG2 as a part of a fully automated system that reliably solves geometry problems from natural language input. Code: \url{https://github.com/google-deepmind/alphageometry2}.
\end{abstract}

\begin{document}

\maketitle

\tableofcontents

\section{Introduction}

Advancing reasoning capabilities of artificial intelligence (AI) systems, especially in tackling complex mathematical problems, has been a decade-long holy grail of AI research. Even with the rapid developments of large language models, most systems until now still struggle with the fundamental understanding of geometric objects (e.g., points, lines, circles) and their interactions. This had motivated our previous work, \ag{}~\cite{Trinh2024}, a neuro-symbolic system capable of solving Euclidean geometry problems at the Olympiad level. Specifically, \ag{}~(AG1) combines the power of a neural language model, that is creative in suggesting insights for tackling challenging problems, with a symbolic engine, that uses a formal language to describe geometric constructs and deduction rules to reliably derive new properties. Trained on 100M synthetically generated examples of diverse levels of complexity, AG1 was able to solve geometry problems at the level of the International Mathematical Olympiad (IMO), a prestigious competition for the world's brightest high-school mathematicians.

Broadly speaking, there are two main approaches for automatically solving geometry problems. The first group is about bashing the geometry problems algebraically with Wu's method~\cite{chou1985proving, wu2008decision}, Area method~\cite{chou1993automated, chou1994machine}, or Gröbner bases~\cite{kapur1986geometry, kapur1986using}. The second group, on the other hand, relies on synthetic techniques, such as Deduction database~\cite{chou2000deductive}, or Full angle method~\cite{chou1996automated}. AG1 uses the latter, as a more human-like approach suitable for transferring the research knowledge to other domains. The neuro-symbolic system that AG1 employed has demonstrated a significant step towards mastering the domain of Euclidean geometry.
However, despite its success, AG1 exhibited limitations in several key areas. Its performance was constrained by the scope of its domain-specific language, the efficiency of the symbolic engine, and the capacity of the initial language model. As a result, when considering all the recent IMO geometry problems from the year 2000 until now, AG1 can only achieve a solving rate of 54\%.

This paper introduces AlphaGeometry2 (AG2), a substantial upgrade that addresses the aforementioned limitations and significantly enhances performance. First, we expand the domain language to encompass a wider range of geometric concepts, including locus theorems with moving constructs and linear equations of angles, ratios, and distances. We also introduce a significantly faster and more robust symbolic engine, incorporating optimizations such as a reduced rule set and enhanced handling of double points. AG2 leverages a more powerful language model based on Gemini~\cite{team2024gemini} and trained on an order of magnitude larger and more diverse dataset. To further improve performance, we develop a novel search algorithm, {\it Shared Knowledge Ensemble of Search Trees (SKEST)}, that explores a broader range of auxiliary construction strategies and employs a knowledge-sharing mechanism to expand and accelerate the search process. Finally, we make progress towards building a fully automated and reliable system that solves geometry problems in natural language. To do this we utilize Gemini to translate problems from natural language into the AlphaGeometry language and implement a new automated diagram generation algorithm.

These enhancements culminate in a substantial improvement in performance. Specifically, AG2 achieves a new state-of-the-art solving rate of 84\% on \textit{all} IMO geometry problems from 2000 to 2024, compared to 54\% achieved in AG1. This demonstrates a significant leap forward in AI's ability to tackle challenging mathematical reasoning tasks, surpassing an average IMO gold medalist. In summary, the key improvements in \ag2 include:
\vspace{-5mm}

\begin{itemize}
\item {\it Expanded domain language:} covering locus-type theorems of moving objects, linear equations of angles/ratios/distances, and non-constructive problem statements.
\item {\it Stronger and faster symbolic engine}: introducing an optimized rule set, better handling of double points, and a two-order-of-magnitude faster implementation in C++.
\item {\it Enhanced Language Model}: leveraging the Gemini architecture, trained on an order of magnitude larger and more diverse dataset.
\item {\it Advanced Novel Search Algorithm}: introducing {\it Shared Knowledge Ensemble of Search Trees (SKEST)} that enables utilizing multiple search trees with knowledge sharing.
\end{itemize}

For related work and a discussion on how the AlphaGeometry2 positions among other systems reported in the literature, we refer the readers to Appendix \ref{app:related-work}.

% ====================================================
% ====================================================
% ====================================================

\section{More general domain language}
\label{sec:domain-lang}

First introduced in \cite{Trinh2024}, AG1 uses a simple domain-specific language that consists of nine basic ``predicates'' listed in Table~\ref{tab:ag1-predicates}. While these predicates are sufficient to cover 66\% of all geometry problems for 2000-2024 IMO, AG1 language does not allow talking about linear equations, movements of points/lines/circles or common computational problems such as ``Find the angle ...'' (e.g., IMO 2009 P4). Below, we explain how AG2 addresses these and other challenges.

\begin{table*}[th!]
\centering
\footnotesize
\begin{tabular}{ll}
\toprule
Name & Meaning \\
\midrule
\code{cong a b c d} & $AB = CD$ \\
\code{perp a b c d} & $AB \perp CD$ \\
\code{para a b c d} & $AB \parallel CD$ \\
\code{coll a b c} & $A, B, C$ are collinear \\
\code{cyclic a b c d} & $A, B, C, D$ are concyclic points \\
\code{eqangle a b c d e f g h} & Directed angle between $AB$ and $CD$ is equal to the one between $EF$ and $GH$ \\
\code{eqratio a b c d e f g h} & $AB/CD=EF/GH$ \\
\code{aconst a b c d x} & Angle between $AB$ and $CD$ is equal to $x$, where $x \in [0, 180)$ \\
\code{rconst a b c d y} & $AB:CD = y$ where $y$ is a constant \\
\bottomrule
\end{tabular}
\caption{AG1 predicates.}
\label{tab:ag1-predicates}
\end{table*}

First of all, AG2 adds two predicates to allow questions of type ``Find x'':

\begin{enumerate}
    \item \code{acompute a b c d} means {\it ``Find the angle between $AB$ and $ CD$''}.
    \item \code{rcompute a b c d} means {\it ``Find the ratio $AB/CD$''}.
\end{enumerate}

In some geometry problems, including the one appearing at IMO 2024, there are linear equations of geometric quantities (angles, distances) that AG1 cannot capture. To express these notions, AG2 adds the following three predicates:

\begin{enumerate}
    \item \code{distmeq a1 b1 a2 b2 ... an bn t1 t2 ... tn y} means $t_1 \log(A_1B_1) + t_2 \log(A_2B_2) + ... + t_n \log(A_nB_n) + y = 0$.
    \item\code{distseq a1 b1 a2 b2 ... an bn t1 t2 ... tn} means $t_1 A_1B_1 + t_2 A_2B_2 + ... + t_n A_nB_n = 0$.
    \item \code{angeq a1 b1 a2 b2 ... an bn t1 t2 ... tn y} means $t_1 d(A_1B_1) + t_2 d(A_2B_2) + ... + t_n d(A_nB_n) + y = 0$ where $d(AB)$ is the angle between the undirected line $AB$ and the horizontal line.
\end{enumerate}

Another category that was not supported in AG1, so-called locus problems, talk about movements of objects such as points, lines, and circles. AG2 captures this through a new predicate syntax. Table~\ref{tab:locus-syntax} lists 11 locus cases with the corresponding predicate and their syntax. Here we make use of one new token \code{*} to serve as the fixed-point placeholder.

\begin{table*}[th!]
\centering
\begin{tabular}{llll}
\toprule
Case & Name & Subcase & Syntax for question \\
\midrule
1 & circle through fixed points & circumcircle a b c & \code{? cyclic a b c * : X} \\
2 & & center a radius b c & \code{? cong b c a * : X} \\
3 & line through fixed points & line a b & \code{? coll a b *: X} \\
4 & & bline of a b & \code{? cong a * b * : X} \\
5 & & pline of a b c & \code{? para b c a * : X} \\
6 & & tline of a b c & \code{? perp b c a * : X} \\
7 & point on fixed line & & \code{? coll a * * : X} \\
8 & point on fixed circle & & \code{? cyclic a * * * : X} \\
9 & fixed distance & & \code{? cong a b * * : X} \\
10 & fixed direction & & \code{? para a b * * : X} \\
11 & fixed angle & & \code{? eqangle a b a c * * * * : X} \\
\bottomrule
\end{tabular}
\caption{The 11 types of locus-type statements, and their corresponding predicate syntax in the AG domain language.}
\label{tab:locus-syntax}
\end{table*}

Furthermore, in AG2 proofs, we include explicit predicates to represent diagram checks for topological/non-degeneracy conditions:

\begin{enumerate}
    \item \code{sameclock a b c d e f} means the direction $A \rightarrow B \rightarrow C$ has the same clockwise direction to $D \rightarrow E \rightarrow F$.
    \item \code{noverlap a b} means $A$ and $B$ are distinct points.
    \item \code{lessthan a b c d} means $AB < CD$, which is a statement used in the SSA triangle congruence theorem.
\end{enumerate}

AG2 can also prove points being non-distinct by introducing a new predicate, \code{overlap a b} (points $A$ and $B$ are overlapping points), where any predicate involving $A$ can also be used for $B$ and vice versa. During the deduction closure, overlapping points can be defined by being a center of the same circle; we therefore introduce another predicate \code{cyclic\_with\_center} to capture this scenario. Here, \code{cyclic\_with\_center a1 a2 ... an x} means $a_1 = a_2 = \cdots = a_x$ is the center of the circle that goes through $a_{x+1} ... a_n$ (in case $x=0$, it is equivalent to \code{cyclic}).

Notice that, when describing a problem, AG1 uses at most 2 predicates to define a point, i.e. each point is defined as the intersection between at most two objects (line or circle). This limits AG1 to only constructive problems - problems where all points can be straightforwardly constructed by following their definition order and taking the intersection of two well-defined objects. In AG2, we relax this constraint to cover more problems where points can be defined by at least three predicates, making the diagram construction non-trivial. Our approach for automating this process is discussed in the next section.

All changes described in this section \textbf{improve the AG domain language coverage from 66 to 88\%} on \textit{all} 2000-2024 IMO geometry problems. The remaining 12\% contain 3D geometry, inequalities, non-linear equations, and countably many points (i.e. problems that have $n$ points where $n$ is an arbitrary positive integer). All problems (covered and not covered) by AG1 and AG2 can be found on Figure~\ref{fig:ag-solved}. Not covered are referred as "Not attempted".

% ====================================================
% ====================================================
% ====================================================

\section{Stronger and faster symbolic engine}
\label{sec:ddar}

A symbolic engine is a core component of AlphaGeometry. We call it DDAR, Deductive Database Arithmetic Reasoning. It is an algorithm to compute the deduction closure, i.e. the set of all deducible facts given a core set of initial facts. DDAR builds this deduction closure by following a fixed set of deduction rules, iteratively adding new facts into the deduction closure until no more can be added.

DDAR drives both the training data generation for our language model and the search for deduction steps during the test-time proof search. In both scenarios, speed is crucial. Faster data generation allows larger and more aggressive data filtering, while faster proof search enables more extensive search, which increases the likelihood of finding a solution within a given time budget. 

There are three main DDAR improvements that will be discussed in the next three sections.
\begin{itemize}
    \item Capability of handling double points.
    \item Faster algorithm.
    \item Faster implementation.
\end{itemize}

\subsection{Handling double points}
\label{subsec:double-points}

While re-implementing DDAR, we tried to keep approximately the same logical strength as the original algorithm, just a little stronger because of the implementation differences (for example Thales Theorem was replaced with a more general Central Angle Theorem). However, DDAR1 is missing one key feature, which is crucial for tackling hard problems: it is unable to accept two points with different names and the same coordinates.

\begin{figure*}[t]
	\centering
	\includegraphics[width=\textwidth]{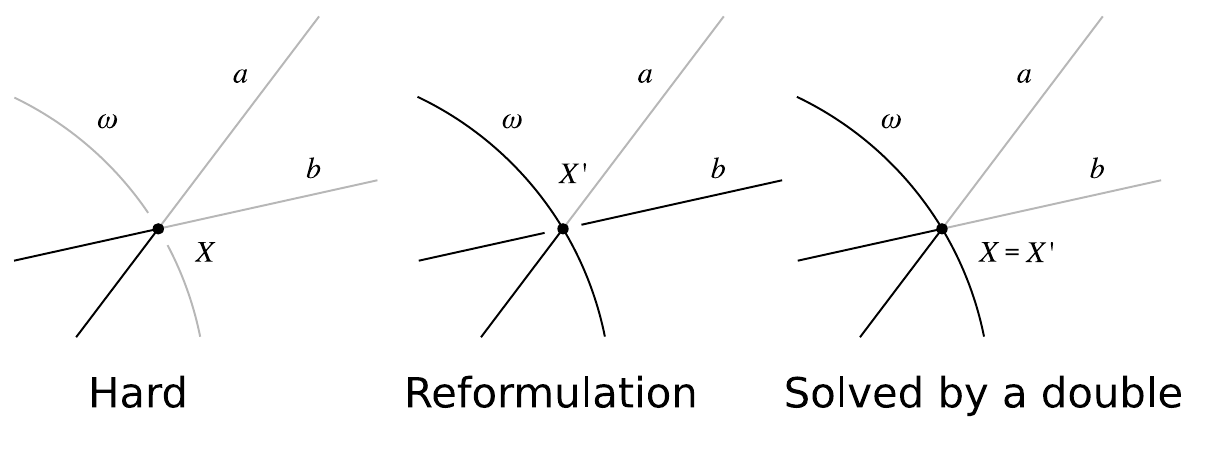}
	\caption{Handling ``double" points in AG2. It is hard to prove that the intersection of $a$, $b$ is on $\omega$. But if a language model suggests a construction $X' \in a \cap \omega$, then DDAR can prove the goal by proving $X' \in b$, and hence $X = X'$.}
	\label{fig:double-point}
\end{figure*}

For example, imagine a problem where we intersect two lines $a, b$ at a point $X$, and intend to prove that $X$ lies on a certain circle $\omega$. The most plausible approach might be via a reformulation – instead of proving that the intersection of $a, b$ is on $\omega$, we prove that the intersection of $a, \omega$ lies on $b$. This is equivalent, yet can be much easier to prove because we can move angles on the circle. For illustration see Figure~\ref{fig:double-point}. To do such ``reformulation'' of reasoning with double points, we proceed through the following 4 steps:
\begin{itemize}
    \item Construct a new point $X'$ as the intersection of $a, \omega$ (we don't know yet that $X'$ coincides with $X$). This is an auxiliary construction that must be predicted by a language model.
    \item Prove that $X$ lies on $b$.
    \item Since both $X$ and $X'$ lie on both, $a, b$, we conclude $X = X'$.
    \item Consequently $X$ lies on $\omega$.
\end{itemize}

\subsection{Faster algorithm}

The DDAR1 algorithm is processing a list of rules and tries to apply each rule to all combinations of points. This process involves a candidate searching step, whose time complexity is polynomial in the number of points, and a clause matching step, whose time complexity is exponential in the number of clauses per premise. In theory, the worst case for searching similar triangle candidates in AG1 is $O(N^8)$, which is one of the most time-consuming steps. The exponential clause matching step is another expensive step. To make the search more efficient, we take all essential rules and hard-code search for their application, which reduces the number of queries for the AR sub-engine to at most cubic. Furthermore, we discard the explicit rules for angles and distances (e.g., about perpendicular or parallel lines) -- all such deductions happen automatically in the AR engine.

The two main time-consuming parts of DDAR are a search for similar triangles and a search for cyclic quadrilaterals. In AG2, we designed an improved DDAR2 algorithm. For similar triangles, we go through all triples of points, hash their ``shape'', and detect a similar pair if the shape is recognized twice. For cyclic quadrilaterals, we go through all pairs (point $X$, segment $AB$), and hash the value of $(A, B, \angle AXB)$. If such a triple repeats, we get a cyclic quadrilateral. By the ``value'' of segment $AB$, or $\angle AXB$, we mean a symbolic normal form calculated by the AR-submodule. This submodule keeps track of known linear equations between angles, distances, and log-distances, understands its algebraic consequences, and can reduce any linear expression to its normal form.

\subsection{Faster implementation}

While the new algorithm already significantly accelerates DDAR, we make further speed improvements by implementing its core computation (Gaussian Elimination) in C++. The new C++ library, which is exported into Python via pybind11 \cite{pybind11}, is over 300 times faster than DDAR1. In order to benchmark the speed improvement, we select a set of 25 IMO problems that cannot be solved by DDAR (see Figure~\ref{fig:ag-solved}), and run the test 50 times on a machine with AMD EPYC 7B13 64-core CPU. While on average DDAR1 finishes its computations in $1179.57 \pm 8.055$ seconds, DDAR2 is much faster - finishing in $3.44711 \pm 0.05476$ seconds\footnote{The average running time may vary depending on the machine status at different times.}.

% ====================================================
% ====================================================
% ====================================================

\section{Better synthetic training data}
\label{sec:data}

Supplementing the symbolic engine with a language model was a key to AG1's success, bringing the solve rate from 14 (pure deduction proofs) to 25 out of 30 selected IMO problems \cite{Trinh2024}. This language model was trained on a large amount of algorithmically generated synthetic data. In AG2, we use the same procedure.

Similar to AG1, our synthetic data generation method starts by sampling a random diagram, and uses the symbolic engine to deduce all possible facts from it. For each of the deduced facts, a traceback algorithm is used to extract the corresponding premises, auxiliary points, and deduction steps that prove the fact. Our data generation method deliberately avoids the use of human-crafted problems as initial diagram seeds, and strictly starts from random diagrams. This design choice eliminates the risk of data contamination and allows for the exploration of theorem distributions that may extend beyond established human knowledge. This approach contrasts with methods like TongGeometry \cite{zhang2024proposing}, which rely on human expertise and existing problem diagrams to guide and filter data generation. In AG2, we keep using random diagrams as initial seeds and continue to push for better synthetic training data.

\begin{figure*}[ht!]
    \begin{subfigure}[b]{\textwidth}
        \centering
        \includegraphics[width=\textwidth]{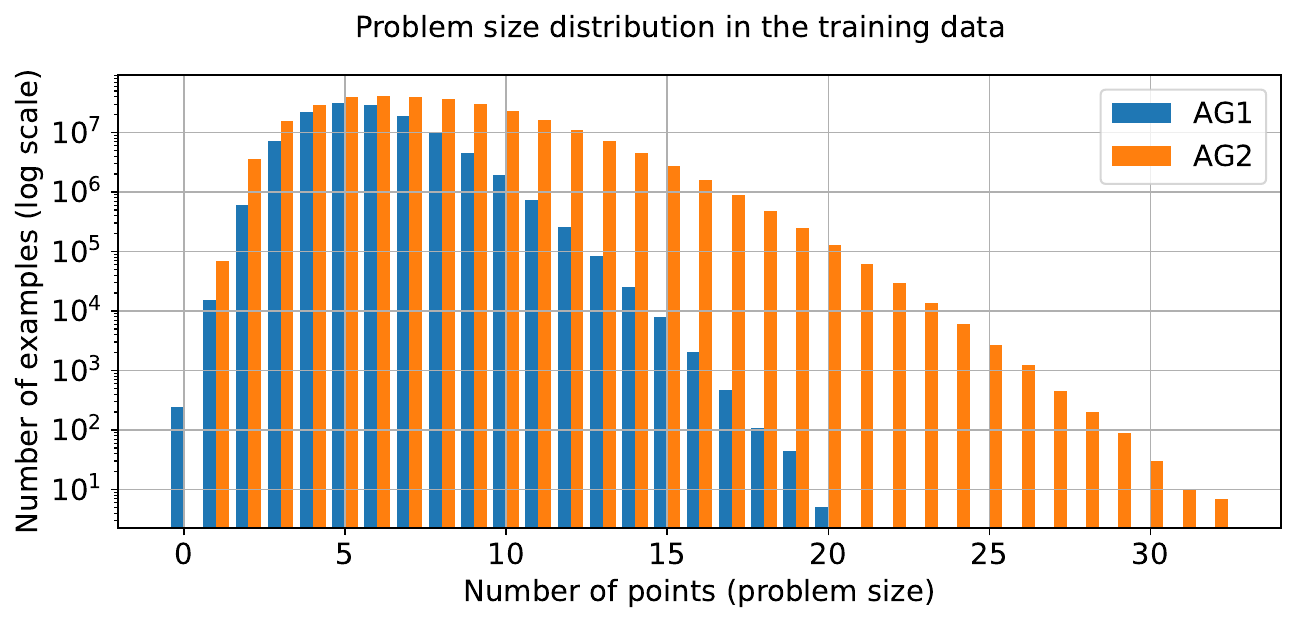}
        \caption{AG2 includes more complicated/longer problems compared to AG1.}
    \end{subfigure}
    \centering
    \begin{subfigure}[b]{0.55\textwidth}
        \centering
        \includegraphics[width=\textwidth]{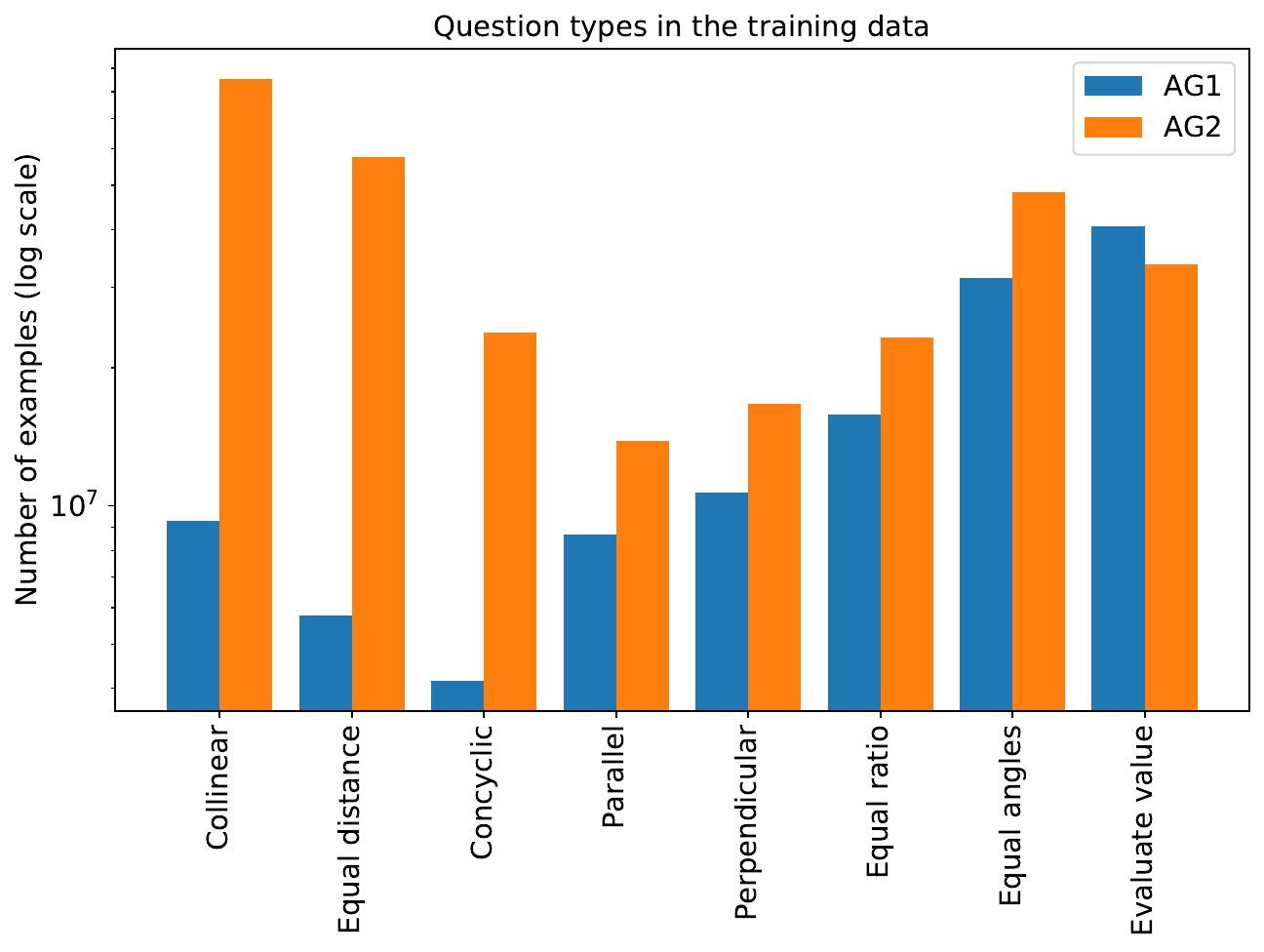}
        \caption{AG2 has a more balanced distribution of examples per question type.}
    \end{subfigure}
    \hfill
    \begin{subfigure}[b]{0.35\textwidth}
        \centering
        \includegraphics[width=\textwidth]{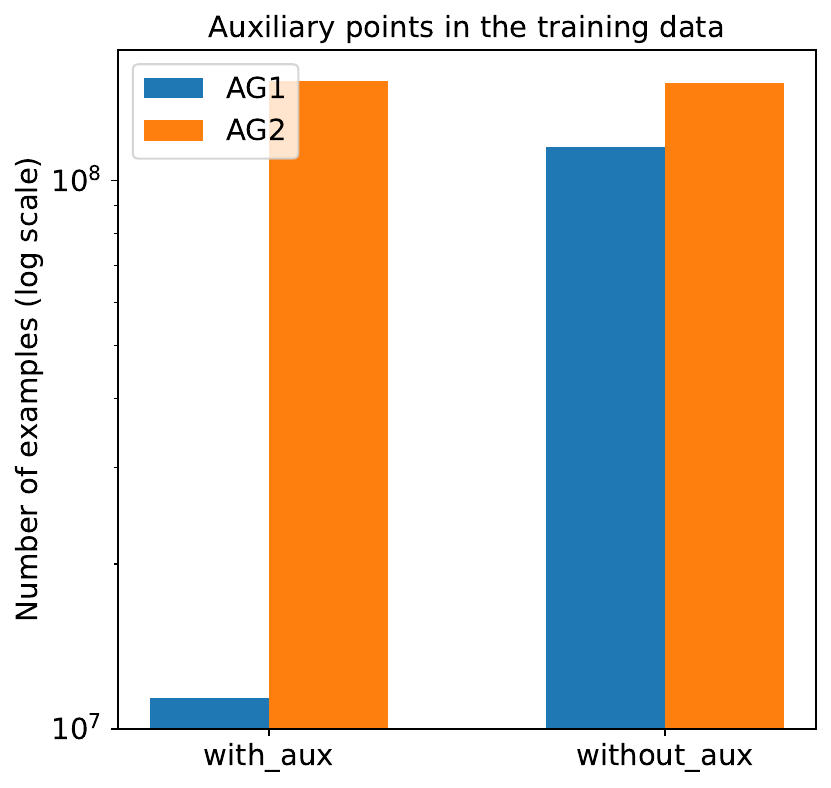}
        \caption{AG2 has a much more balanced mix between proofs with auxiliary points and proofs without (50:50 in AG2 vs 9:91 in AG1).}
    \end{subfigure}
    \caption{Training data distributions for AG2 compared to AG1.}
    \label{fig:ag2-data}
\end{figure*}

\paragraph{Larger, more complex diagrams and better data distribution.} First of all, we scale up resources for data generation, and do more careful re-balancing of the data distribution. As demonstrated on Figure~\ref{fig:ag2-data}, compared to AG1, AG2:

\begin{itemize}
    \item Explores random diagrams at twice the size, allowing for extracting much more complex problems.
    \item Produces theorems at up to 2x more complex, i.e. number of points and premises.
    \item Produces up to 10x more complex proofs, i.e. 10x more proof steps.
    \item Has a more balanced data distribution between question types.
    \item Has a more balanced data distribution between problems with and without auxiliary points.
\end{itemize}

\paragraph{More types of theorems.} Besides generating theorems that prove classic statements such as ``AB = CD'', AG2 data generating algorithm also produces problems of ``locus'' type, i.e. asserting statements such as ``When X moves on line/circle Y, then Z moves on a fixed line/circle T''. Introduced in Section~\ref{sec:domain-lang}, these statements are not supported in the AG1 data generation algorithm, as there is no notion of movement and movement dependency. In AG2, we record the movement dependency for each point $X$ during random diagram generation through a function $P(.)$ with the following definition:

$P(A)$: the set of points that control the movements of $A$, where $A$ is a point or a set of points, defined in a constructive problem statement. Two examples of $P$ are presented in Table \ref{tab:movement-examples} and all cases where locus-type statements are detected are shown in Table \ref{tab:locus-cases}.

\begin{table*}[ht!]
\centering
\begin{tabular}{l|l}
\textbf{If} & \textbf{Then} \\
\hline
\code{a = midpoint b c, d = midpoint a c} & $P(d) = \{b, c\}$ \\
\code{a = on\_line b c} & $P(a) = \{a, b, c\}$ \\
\end{tabular}
\caption{Two examples of $P$. Top row: since $d$ is uniquely defined as the midpoint of $a$ and $c$, and $a$ is uniquely defined as the midpoint of $b$ and $c$, the source of movement for $d$ is $b$ and $c$. Second row: Since $a$ can be anywhere on line $bc$, $a$ itself is also a part of its movement source.}
\label{tab:movement-examples}
\end{table*}

\paragraph{Faster data generation algorithm.} We also improved the speed of the data generation algorithm. Recall that in AG1, we first run deduction closure on random diagrams, and ``traceback" to obtain the minimal problem and minimal proof that proves each fact in the closure. To obtain the minimal problem in AG1, we have to exhaustively remove different subsets of points from the problem and rerun DDAR to check provability. Such a search can find the subset with the smallest cardinality, however, as an exponential search, it is infeasible for a larger number of points. Therefore, we switched to a greedily discarding algorithm shown in Figure~\ref{fig:greedy-alg}, which uses just a linear number of checks for whether a set of points suffices to prove the goal. The greedy algorithm is guaranteed to find a minimal set of points with respect to inclusion as long as the check is monotonic (if $A \subseteq B$, then $\texttt{check\_provable}(A) \Rightarrow \texttt{check\_provable}(B)$). In reality, we also require the pruned set to remain closed under construction dependencies (so that we can still run a random construction). If we incorporate this condition into the \texttt{check\_provable} predicate, it stops being monotonic. This difficulty can be fixed by processing the points via the algorithm from Figure~\ref{fig:greedy-alg} in a reverse-topological order (first points that do not depend on any other points, and last the initial points of the construction).

\begin{figure}[ht!]
\begin{lstlisting}[language=Python]
def prune_points(
        points : set[Point],
        check_provable: Callable[[set[Point]], bool]):
    pruned = set(points)
    for p in reverse_topological(points):
        if check_provable(pruned - {p}):
            pruned = pruned - {p}
    return
\end{lstlisting}
\caption{Basic greedy algorithm to find a minimal set of points satisfying a monotonic predicate \texttt{check}.}
\label{fig:greedy-alg}
\end{figure}

% ====================================================
% ====================================================
% ====================================================

\section{Novel search algorithm}
\label{sec:stronger-search-algorithm}

\begin{figure*}[ht!]
    \centering
    \includegraphics[width=\textwidth]{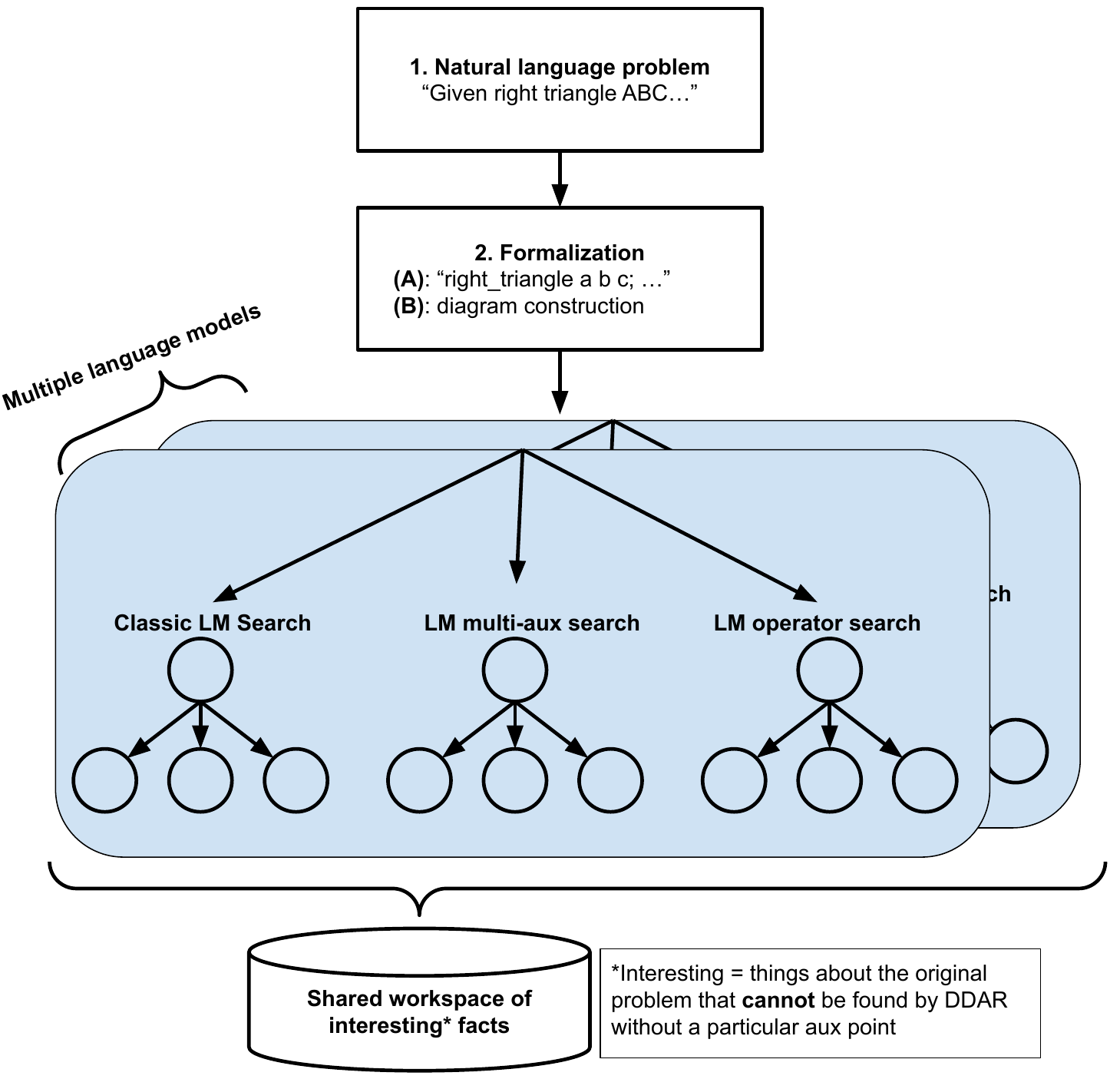}
    \caption{Overview of our search algorithm. We employ several different search trees which can share facts they proved via a special knowledge sharing mechanism.}
    \label{fig:search-algo}
\end{figure*}

In AG1, we use a simple beam search to discover proofs. In AG2, we design a novel search algorithm, in which several differently configured beam searches are executed in parallel and are allowed to help each other through a knowledge sharing mechanism (see Figure~\ref{fig:search-algo}). To improve the robustness of our system, we use multiple different language models for each search tree configuration. We call this search algorithm Shared Knowledge Ensemble of Search Trees (SKEST).

It works as follows. In each search tree, a node corresponds to one attempt at auxiliary construction followed by one attempt of the symbolic engine run. If the attempt succeeds, all search trees terminate. If the attempt fails, the node will write down the facts that the symbolic engine managed to prove into a shared facts database. These shared facts are filtered such that they are not the auxiliary point specific to the node itself, but only relevant to the original problem. This way, these facts can also be useful to the other nodes in the same search tree, and the nodes in different search trees. Below we list various types of search trees, which we employ to make sure different parts of the search space are explored effectively:

\begin{itemize}
    \item ``Classic" search tree: the same beam tree search used in AG1, where a language model is asked to produce one auxiliary point at each node.
    \item Tree predicting multiple auxiliary points at each node: a language model is allowed to produce as many auxiliary points as it wants at each tree node. Recall that this is possible because our LM is trained to produce full proofs, starting with auxiliary points and followed by the deduction steps\footnote{See a more detailed discussion on producing full proofs with a language model alone in Section~\ref{app:full-proof}}. Note that even though we want our models to generate all necessary auxiliary points in one query, in practice, we observe the need to call the model multiple times given previously produced auxiliary points. Allowing the model to produce multiple auxiliary points accelerates finding a solution and effectively increases the tree search depth.
    \item Tree predicting different types of aux points uniformly. Recall that LM outputs for auxiliary points look like \code{x00 a : cong a b c d (00) coll a e f (01)}, i.e. ``construct point a such that a b = c d and a e f are collinear". Typically, to predict aux points, we prompt the language model with the first token \code{x00} and let it generate the rest. Here, instead, we prompt the LM with \code{x00 a : cong}, \code{x00 a : coll}, \code{x00 a : cyclic}, \code{x00 a : perp}, etc. to force uniform distribution across the first 4 tokens, and then let the LM generate the rest.
    \item Deep but narrow tree (e.g. beam size 64 and depth 10).
    \item Shallow but wide tree (e.g. beam size 512 and depth 4).
\end{itemize}

\paragraph{System design details.} For proof search, we use TPUv4 to serve multiple replicas per model\footnote{The exact number of TPUs depends on the model size.} and let different search trees within the same model to query the same server under their own search strategy. Besides running these search trees asynchronously, we also run the LM workers asynchronously with DDAR workers. The LM workers write down the content of the nodes they explored to a database, and the DDAR workers asynchronously pick up these nodes and attempt them. The DDAR workers coordinate between themselves to make sure they divide work equally. A single DDAR worker pool is shared across different problems (if multiple problems are solved at once), such that problems that got solved earlier release its own DDAR compute resources for the rest of the problems that are being solved.

% ====================================================
% ====================================================
% ====================================================

\section{Better language model}
\label{sec:lm}

The final AG2 improvement is a new language model. In this section we discuss our new training and inference setups.

\subsection{Training setup}
\label{sec:training-setup}

AG1 language model, a custom transformer, was trained in an unsupervised fashion in two phases: training on problems with and without auxiliary constructions followed by training on only problems that contain auxiliary constructions. For AG2, we leverage the Gemini training pipeline and simplify training to just one phase: unsupervised learning on all data. Our new language model is a sparse mixture-of-expert Transformer-based model that builds on Gemini \cite{team2024gemini} and is trained on AG2 data described in Section~\ref{sec:data}. We train multiple models of different sizes using three training setups:
\begin{enumerate}
    \item Training from scratch with a custom tokenizer in the domain-specific language (AG1 setup).
    \item Fine-tuning already pre-trained custom math specialized Gemini models in natural language (for more details see Appendix \ref{app:fine-tuning}).
    \item Multimodal training from scratch with an additional image input - a diagram of the given geometry problem (for more details see Appendix \ref{app:multimodal}).
\end{enumerate}
 
Apart from a large synthetic training set of around 300 million theorems, we create three evaluation sets:
\begin{enumerate}
\item Synthetic problem set with and without auxiliary points, ``eval".
\item Synthetic problem set with only auxiliary points, ``eval\_aux".
\item Special set of geometry problems from IMO 2000-2024 that have been solved by AlphaGeometry previously, ``imo\_eval".
\end{enumerate}

All these sets contain full proofs, and we compute perplexity loss on them during training. Note, however, that these are only proxy metrics for two reasons. First, during inference (just like in AG1), we only use auxiliary points suggested by the language model, while the perplexity is computed on the entire proof. Second, there might be multiple ways to solve a given problem, but perplexity is computed for one particular solution. Just like in AG1, our main downstream metric is the solve rate on IMO problems, where the language model produces auxiliary points followed by a DDAR run via beam search described in section \ref{sec:stronger-search-algorithm}. These results will be discussed in Section~\ref{sec:results}.

\begin{figure*}[ht!]
    \centering
    \includegraphics[width=\textwidth]{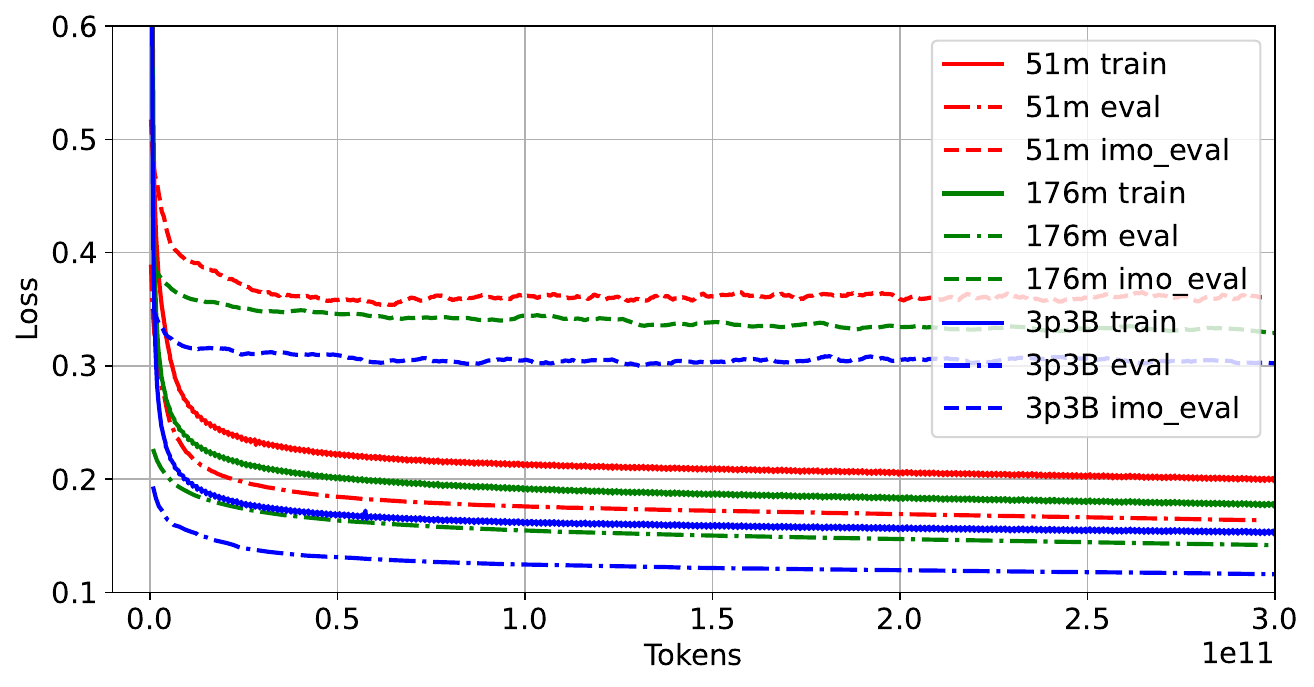}
    \caption{Learning curves for AlphaGeometry2 language models of different sizes in terms of parameter count (``m" - million, ``B" - billion, for example, ``3p3B'' means a model with 3.3 billion parameters). Increasing the model size results in decreasing loss for train, eval and IMO evaluation sets.}
    \label{fig:ag-lang-learning-curves}
\end{figure*}

We train our models with the largest possible batch size allowed by the hardware\footnote{We did not observe any training issues compared to smaller batches.} using TPUv4. A learning rate schedule is a linear warm-up followed by the cosine anneal. Learning rate hyperparameters are determined from scaling laws. On Figure~\ref{fig:ag-lang-learning-curves}, we illustrate learning curves for different-sized Gemini models in terms of parameter count. As expected, increasing the model size decreases perplexity loss for train, eval and our special IMO evaluation set.

\subsection{Inference setup}

\begin{figure*}[ht!]
    \centering
        \begin{minipage}{0.48\textwidth}
        \includegraphics[width=\textwidth]{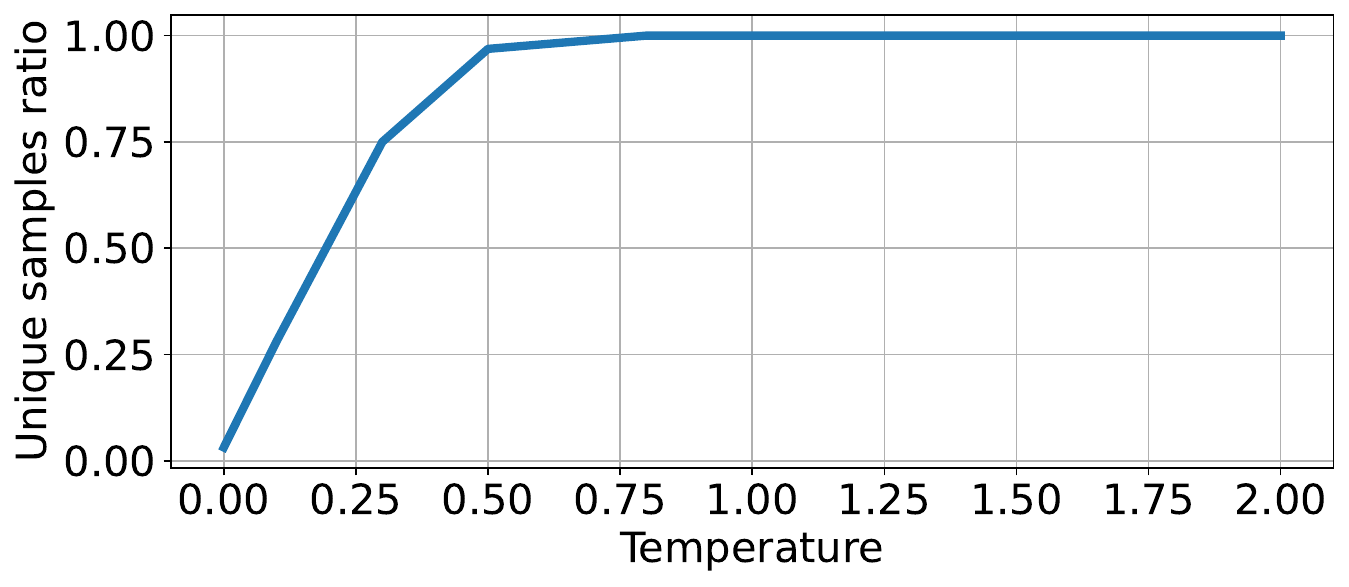}
        \caption{Ratio of unique samples for various temperatures for top-k sampling.}
        \label{fig:temp-vs-num-samples}
    \end{minipage}\hfill
    \begin{minipage}{0.48\textwidth}
        \centering
        \includegraphics[width=\textwidth]{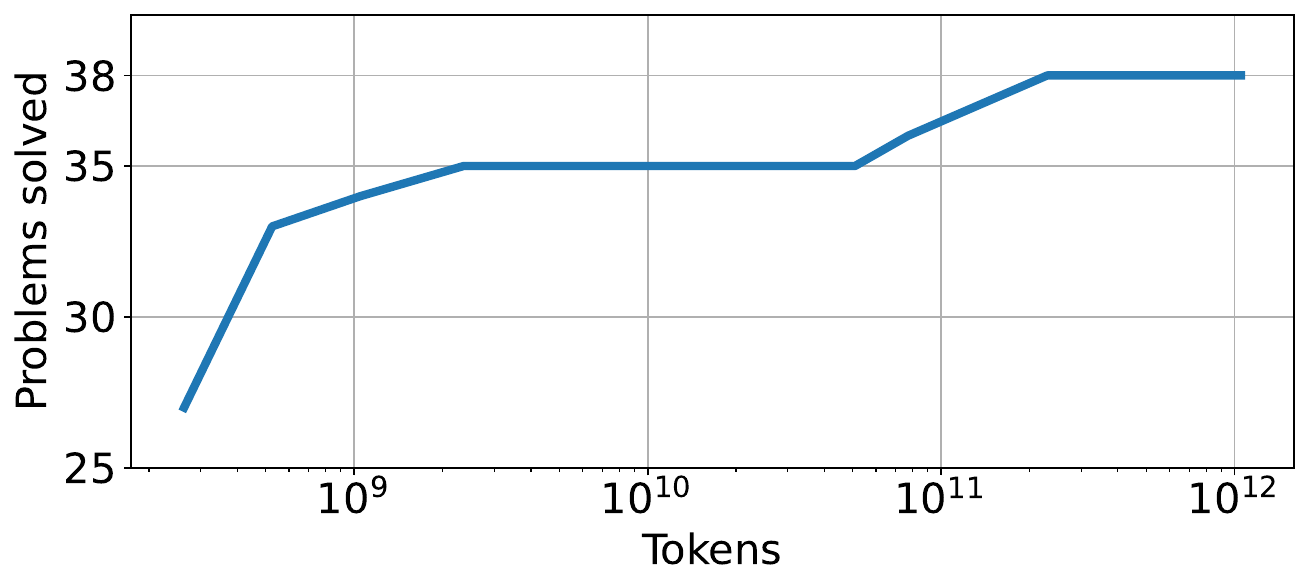}
        \caption{Number of 2000-2024 IMO problems solved by one language model as a function of seen tokens during training.}
        \label{fig:imo-solve-rate-vs-train-step}
    \end{minipage}
\end{figure*}

A new problem is solved via the search algorithm described in section \ref{sec:stronger-search-algorithm} with multiple search trees and multiple language models of different sizes. In contrast to AG1, we use top-k sampling with temperature $t=1.0$ and $k=32$. Note that a high temperature and multiple samples are essential for solving IMO problems. With the greedy decoding $t=0.0, k=1$, and no tree search, our models can solve only two problems out of 26 that require auxiliary constructions. Increasing the temperature to $t=1.0$ and using $k=32$ samples (without a search tree) allows our language models to solve 9 out of 26 problems. Lower temperatures $t < 1.0$ do not produce diverse enough auxiliary constructions (see Figure~\ref{fig:temp-vs-num-samples}), while higher temperatures result in an increasing number LM outputs with a wrong domain language syntax.

\paragraph{The analysis string.} \label{sec:analysis}

In AG1, the interface between LM and DDAR is minimal: DDAR takes auxiliary constructions proposed by LM, and the LM stops proposing auxiliary constructions when DDAR succeeds in finding a solution. In AG2, we enrich this neuro-symbolic interface by letting the LM know about the deductions made by DDAR before proposing auxiliary constructions. Namely, we feed the following information into the LM:

\begin{itemize}
    \item $S_1$: Set of facts deducible by DDAR given the original problem premises.
    \item $S_2$: Set of facts deducible by DDAR given the original problem premises and assuming the goal predicate is also true.
    \item $S_3$: Set of facts that is correct numerically (by inspecting the diagram).
\end{itemize}

Note that by definition, $S_1 \subset S_2 \subset S_3$. Once these three sets are computed, we serialize and concatenate them into a string called {\it analysis string}, using our domain-specific language. This string is fed into the LM, together with the original problem statement as follows: \code{<problem\_statement> serialized($S_1$) serialized($S_2 - S_1$) serialized($S_3 - S_2$)}. In contrast, the input to the AG1 LM is simply \code{<problem\_statement>}.

% ====================================================
% ====================================================
% ====================================================

\section{Results}
\label{sec:results}

\begin{figure*}[ht!]
    \centering
    \includegraphics[width=\textwidth]{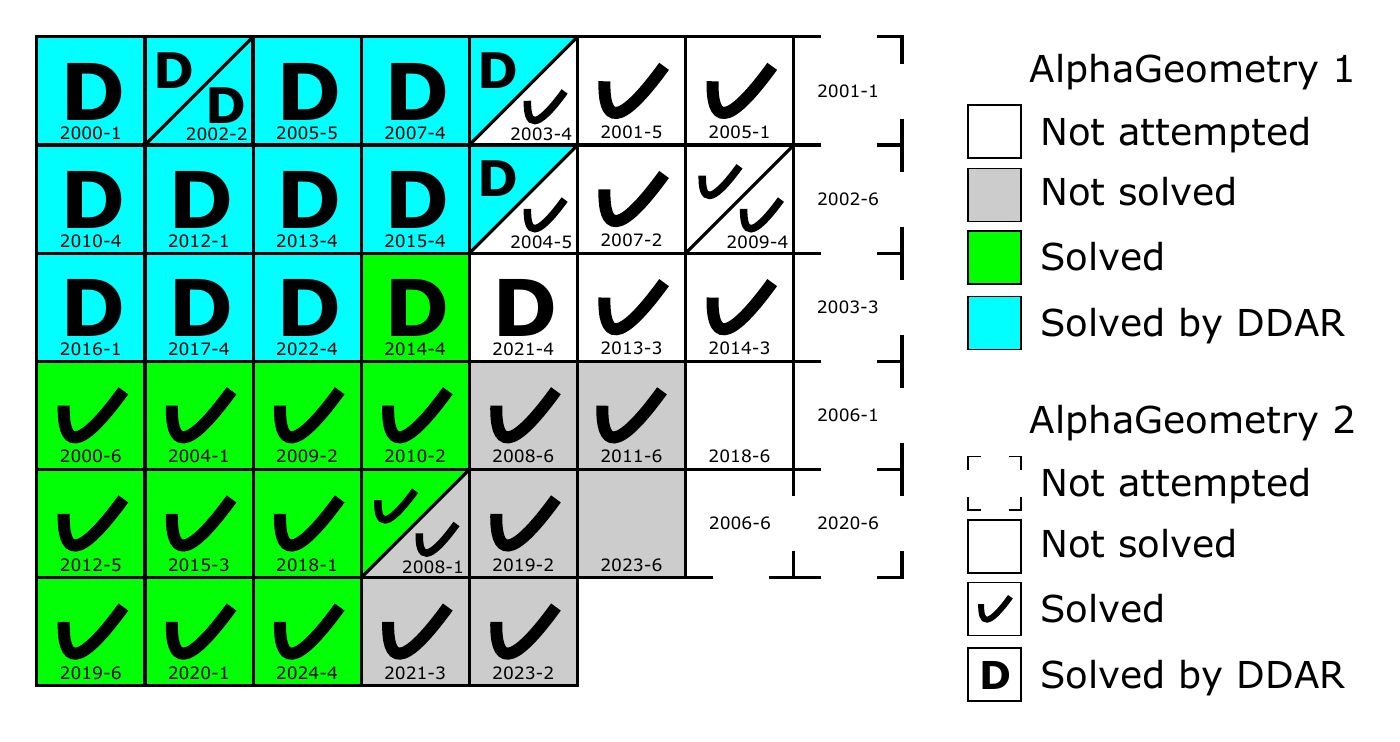}
    \caption{AlphaGeometry2 results on all 2000-2024 IMO geometry problems. Problems are grouped together based on their status, and ordered chronologically within the groups.}
    \label{fig:ag-solved}
\end{figure*}

Our main downstream metric is the solve rate on IMO geometry problems. There are a total of 45 geometry problems in 2000-2024 IMO, which we translate into 50 AlphaGeometry problems (we call this set IMO-AG-50). Some problems are split into two due to the specifics of our formalization. Figure~\ref{fig:ag-solved} demonstrates our main result: \textbf{AlphaGeometry2 solves 42 out of 50 of all 2000-2024 IMO geometry problems, thus surpassing an average gold medallist for the first time}\footnote{\cite{sinha2024wusmethodboostsymbolic} previously claimed to achieve a gold medalist performance, but it was done on a subset of IMO problems.}. More details are presented in Table~\ref{table:post-imo-2000-results}, which compares various AG2 configurations with other systems, such as AG1 \cite{Trinh2024} and TongGeometry \cite{zhang2024proposing}. We also perform an additional evaluation on a new set of 30 hardest IMO shortlist problems, which are formalizable in the AG2 language, and which have never appeared at IMO. For these additional results, see Appendix \ref{app:imosl-results}.

\begin{table}[ht!]
    \centering
    \begin{tabular}{|l|l|l|}
    \hline
        System description & IMO-AG-50 solved & IMO-AG-30 solved \\
        \hline 
        OpenAI o1 & 0 & 0 \\
        Gemini thinking & 0 & 0 \\
        AG1 DDAR \cite{Trinh2024} & 14 & 14 \\
        AG2 DDAR & 16 & 15 \\
        TongGeometry DD \cite{zhang2024proposing} & - & 18 \\
        Average bronze medalist & 27.1 & 19.3 \\
        Wu with AG1 DDAR \cite{sinha2024wusmethodboostsymbolic} & - & 21 \\
        Average silver medalist & 33.9 & 22.9 \\
        AG1 \cite{Trinh2024} & 27 & 25 \\
        Average gold medalist & 40.9 & 25.9 \\
        Wu + AG1 \cite{sinha2024wusmethodboostsymbolic} & - & 27 \\
        TongGeometry w/o value \cite{zhang2024proposing} & - & 28 \\
        AG2 with AG1 setup (a single search tree) & 38 & 28 \\
        TongGeometry full setting \cite{zhang2024proposing} & - & \textbf{30} \\
        AG2 full setting (multiple search trees) & \textbf{42} & \textbf{30} \\
    \hline
    \end{tabular}
    \caption{Evaluation on IMO-AG-50 benchmark. IMO-AG-50 contains \textit{all} IMO 2000-2024 geometry problems, while IMO-AG-30 introduced in \cite{Trinh2024} contains only a subset formalizable in terms of the AG1 language.}
    \label{table:post-imo-2000-results}
\end{table}

\begin{figure*}[ht!]
    \centering
    \includegraphics[width=\textwidth]{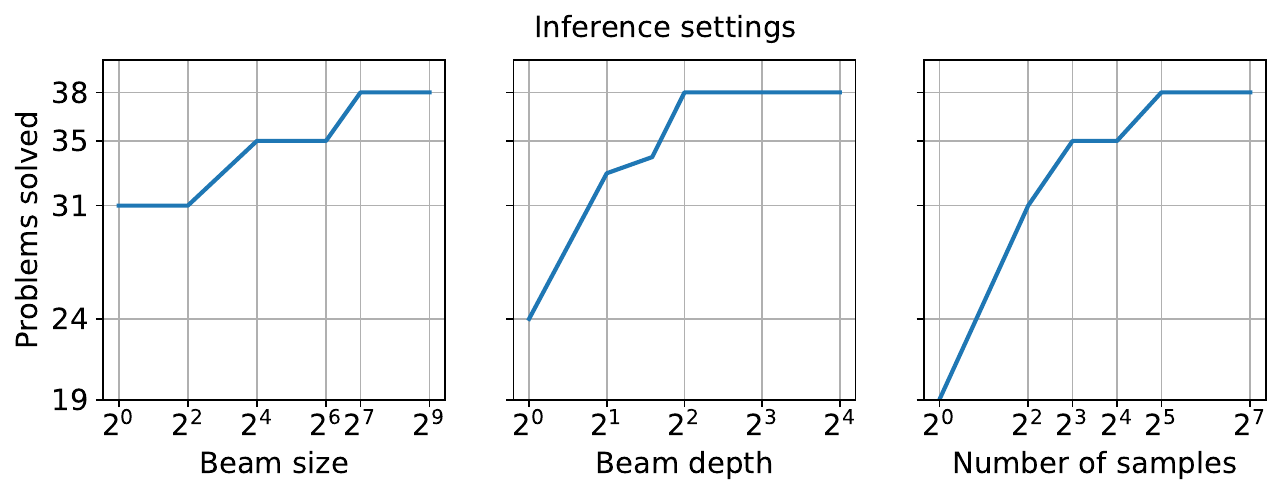}
    \caption{Number of 2000-2024 IMO geometry problems solved for different inference settings with one search tree. We start with beam size 512, beam depth 4, 32 samples and vary one of the parameters while keeping others fixed.}
    \label{fig:inference-ablations}
\end{figure*}

On Figure~\ref{fig:imo-solve-rate-vs-train-step}, we present the IMO solve rate as a function of training time (seen tokens during training) for one language model coupled with DDAR via the "classical" tree search described in Section~\ref{sec:stronger-search-algorithm}. Interestingly, AlphaGeometry2 can already solve 27 out of 50 problems after only 250 training steps with a batch size of 256, or around 200 million tokens\footnote{Note that even without the language model, AlphaGeometry2 can solve 16 problems with its symbolic engine alone (see Figure~\ref{fig:ag-solved}).}. We also run ablation studies on how inference settings affect the overall performance (see Figure~\ref{fig:inference-ablations}). For a single search tree, we find that the optimal configuration is a beam size of 128, a beam depth of 4, and 32 samples. More samples or a larger beam search do not help solve more problems. 

Our geometry experts and IMO medalists consider many AlphaGeometry solutions to exhibit \textit{superhuman creativity}. In Appendix \ref{app:ag2-examples} we provide several such examples with the detailed analysis. Out of the unsolved IMO problems, we have 2 attempted but not solved and 6 unformalizable problems. The unformalizable problems involve inequalities and variable number of points, which are currently not covered by the AlphaGeometry2 language. Two of the remaining unsolved IMO problems (IMO 2018 P6, IMO 2023 P6) involve advanced geometry problem-solving techniques such as inversion, projective geometry or radical axis, which are not implemented in our current DDAR. While such problems in theory can be solved without these techniques, such solutions would require longer inference time, longer proofs and more auxiliary constructions to make up for the lack of the aforementioned machinery in our current DDAR, which hinders AlphaGeometry's current problem-solving capabilities.

% ====================================================
% ====================================================
% ====================================================

\section{Conclusions and Future Work}
\label{sec:conclusion}

This paper introduced AlphaGeometry2, a significant upgrade to AlphaGeometry that addresses previous limitations and enhances performance in several key areas. AG2 incorporates a more powerful language model trained on a larger and more diverse dataset, a faster and more general symbolic engine, an expanded domain language, and a novel proof search algorithm. These improvements have resulted in a substantial leap in performance, with AG2 achieving an 84\% solve rate on 2000-2024 IMO geometry problems, significantly improving upon the 54\% solve rate achieved by its predecessor.

We also present several studies related to language models in general. First, in Appendix \ref{app:fine-tuning} we demonstrate that for AlphaGeometry neither a tokenizer, nor a domain language used to train the model, plays a decisive role. We obtained similar results for custom tokenizers with small vocabularies and the generic large Gemini tokenizer. Training in the domain-specific language achieves similar results compared to training in natural language. Second, in the same appendix, we compare training from scratch and fine-tuning language models pre-trained on math datasets. We find that despite training on the same AlphaGeometry dataset, these models learn slightly different skills, and combining them into our novel search algorithm, Shared Knowledge Ensemble of Search Trees, improves the overall solve rate. Third, in Appendix \ref{app:full-proof}, we show that our models are quite capable in generating not only auxiliary constructions, but also full proofs, demonstrating a potential for modern language models to operate without external tools, such as symbolic engines. Finally, in Appendix \ref{app:diagram-generation} we show that AlphaGeometry2 can be used to build a fully automated geometry problem-solving system that takes inputs in natural language and reliably outputs corresponding diagrams and full solutions without any hallucinations.

Despite achieving an impressive 84\% solve rate on all 2000-2024 IMO geometry problems, there is still room for improvement. First, our domain language does not allow talking about variable number of points, non-linear equations, and problems involving inequalities, which must be addressed in order to fully ``solve the Euclidean geometry''. Extending AlphaGeometry to encompass these topics is a substantial undertaking that falls beyond the scope of this work. To illustrate the complexity involved, in Appendix \ref{app:inequality-rules} we present the list of inequality rules the symbolic engine would need to incorporate. Second, AG2 has not solved all IMO and IMOSL problems. We hypothesize that breaking problems into subproblems and applying Reinforcement learning approaches could close this gap.

% ====================================================
% ====================================================
% ====================================================

\section*{Code availability}

Code for the Python implementation of the symbolic engine (DDAR2), along with multiple examples of proven IMO problems, will be shared at \url{https://github.com/google-deepmind/alphageometry2}. Code for training Gemini models is not provided because it uses internal Google infrastructure. Code for running the full tree search with multiple Gemini models is not provided because it relies on internal Google infrastructure and running many workers in parallel.

% ====================================================
% ====================================================
% ====================================================

\section*{Data availability}

We share 27 IMO problems translated to the AlphaGeometry language along with their diagrams and solutions. They can be found in the file \code{test.py} within the provided repository.

% ====================================================
% ====================================================
% ====================================================

\section*{Acknowledgments}
Special thanks to Dawsen Hwang, Edward Lockhart, and Steven Creech for contributions to the development of AlphaGeometry2. We would also like to thank Yifeng Lu, Henryk Michalewski, Ed Chi, David Silver, Pushmeet Kohli, and Demis Hassabis for their thoughtful discussions and support.

\newpage

\appendix

\vskip 0.2in
\bibliography{main}

% ====================================================
% ====================================================
% ====================================================
% ====================================================
% ====================================================
% ====================================================

\begin{table*}[ht!]
\centering
\rotatebox{90}{
\small
\begin{tabular}{p{4cm}p{3cm}p{6cm}p{3cm}c}
\toprule
Given a random diagram, if DDAR proves: & Then let X = & Then if X is nonempty, we create a synthetic proof that says: "when X moves, ... & Auxiliary constructions will be the following points and everything else they depend on & Case \\
\midrule
\code{cong a b c d} & $P(b, c, d) - P(a)$ & circle center $b$, radius $cd$ goes through a fixed point & $a$ & 2 \\
& $P(a) - P(b, c, d)$ & $a$ moves on a fixed circle & $b, c, d$ & 8 \\
& $P(a, b) - P(c, d)$ & the distance between $a$ and $b$ is fixed & $c, d$ & 9 \\
\midrule
\code{cong a b a c} & $P(a) - P(b, c)$ & $a$ moves on a fixed line & $b, c$ & 7 \\
& $P(b, c) - P(a) \geq M$ & $b$ \& $c$ are equidistant to a fixed point, when M move & $a$ & 4 \\
\midrule
\code{cyclic a b c d} & $P(b, c, d) - P(a)$ & the circumcircle of $b$ $c$ $d$ moves through a fixed point & $a$ & 1 \\
& $P(d) - P(a, b, c)$ & $d$ moves on a fixed circle & $a, b, c$ & 8 \\
\midrule
\code{coll a b c} & $P(b, c) - P(a)$ & line $b$ $c$ goes through a fixed point & $a$ & 3 \\
& $P(c) - P(a, b)$ & $c$ moves on a fixed line & $a, b$ & 7 \\
\midrule
\code{eqangle b a b c e d e f} & $P(a, b, c) - P(d, e, f)$ & the angle $a$ $b$ $c$ has a fixed value &  $d, e, f$ & 11 \\
& $P(b) - P(a, c, d, e, f)$ & the point $b$ moves on a fixed circle & $a, c, d, e, f$ & 8 \\
\midrule
\code{para a b c d} & $P(b, c, d) - P(a)$ & The line through $b$ and $\parallel cd$ moves through a fixed point & $a$ & 5 \\
& $P(c, d) - P(a, b)$ & The line $cd$ is always parallel to a fixed line & $a, b$ & 10 \\
& $P(a) - P(b, c, d)$ & $a$ moves on a fixed line & $b, c, d$ & 7 \\
\midrule
\code{perp a b c d} & $P(b, c, d) - P(a)$ & the line through $b$ and $\perp cd$ moves through fixed point & $a$ & 6 \\
& $P(c, d) - P(a, b)$ & the line $cd$ is always $\perp$ to a fixed line & $a, b$ & 10 \\
& $P(a) - P(b, c, d)$ & $a$ moves on a fixed line & $b, c, d$ & 7 \\
\bottomrule
\end{tabular}
}
\caption{17 cases where locus-type statements are detected during data generation. These 17 cases produce 11 different types of locus-type statements, as numbered in the last column.}
\label{tab:locus-cases}
\end{table*}

\newpage

% ====================================================
% ====================================================
% ====================================================

\section{Related work}
\label{app:related-work}

\textbf{Neuro-symbolic system:} At the high level, AlphaGeometry is a neuro-symbolic system with a generative component (i.e. language model). This framework was motivated from the fact that a major limitation of automated theorem provers is the ability to generate original mathematical terms, which could be addressed using language models \cite{polu2020generativelanguagemodelingautomated}. Such framework was used in several papers such as \cite{Trinh2024, wei2024provingineq,kaiyu2025inequalities} to great success. Additionally, because the symbolic system is able to provide consistent and verifiable feedback, neuro-symbolic systems have been augmented with reinforcement learning techniques to give AI agents the ability to self-improve without human intervention \cite{jha2024rlsfreinforcementlearningsymbolic, peng-etal-2023-geodrl}. In contrast, our AG2 does not use any reinforcement learning techniques to achieve gold-medalist level performance on IMO geometry problems.\\
\\
\textbf{LLM-based feedback}: In the era of large language models (LLMs), one might think of using them (instead of symbolic engines) as verifiers due to their versatility. This approach is used in \cite{shinn2023reflexionverbalrl, madaan2023selfrefine, chen2024teachingllmselfdebug, lightman2024letsverify, zhang2025generativeverifiers}. However, this approach lies on the assumption that LLMs are reliable verifiers and this notion has been challenged \cite{huang2024llmcannotselfcorrect, stechly2025onselfverificationlimitation}. Solving Olympiad geometry problems involves performing numerous precise algebraic manipulations consistently throughout a potentially long solution, whereas LLMs are notoriously unreliable even for basic arithmetics \cite{yan2025phdlevelllmstrulygrasparithmetic}. And so, it is natural for AlphaGeometry to favor symbolic engines over LLMs to provide consistent feedback to the solver.\\
\\
\textbf{Visual reasoning}: While AlphaGeometry can solve difficult Olympiad geometry problems and surpass the performance of top human experts, many papers have demonstrated the limitations of various foundation models in performing geometric reasoning \cite{mouselinos-etal-2024-geometric-reasoning-gap, wang2025dollmunderstandgeometry}. As visual thinkers \cite{Tversky2009-think-with-sketch}, humans want to teach AI agents to leverage visual information to further unlock its reasoning capabilities \cite{hu2024visualsketchpad}. Even though AlphaGeometry mostly relies on algebraic reasoning, we discuss incorporating visual modality in Appendix \ref{app:multimodal}.

\section{Fine-tuning of math specialized language models on AG data}
\label{app:fine-tuning}

Even though during the initial transition to AG2, we maintained the AG1 training setup (training from scratch using a custom tokenizer in the AG domain-specific language), it is natural to ask whether fine-tuning language models that already possess problem-solving capabilities can improve the performance. Such fine-tuning is not immediately possible due to the difference in the utilized tokenizers and training language. In this section, we explore the role of the custom tokenizer and the domain-specific language, followed by a discussion about fine-tuning of a math-specialized Gemini model on AG data.

\paragraph{Tokenizers.} Tokenizer is an essential part of modern language models and, more broadly, any foundation models\footnote{Tokenizer-free models is an active area of research, for example, see \cite{deiseroth2024tfreetokenizerfreegenerativellms}}. It is generally believed that a tokenizer might be the major bottleneck in the models' abilities to do math, e.g. see \cite{singh2024tokenizationcountsimpacttokenization}. We investigate this hypothesis in the controlled setting of AlphaGeometry. To do so, we train models of the same architecture with different tokenizers: custom tokenizers with vocabularies of a few thousand tokens and the large language model tokenizers with a vocabulary of ~300k tokens. Recall that our custom tokenizers are created at word-level, i.e. each token has full meaning, as opposed to subword-level tokens. In AG language, there are the following types of tokens:
\begin{enumerate}
    \item Point names: ``a'', ``b'', ``c'', \dots ``z'', ``a1'', \dots, ``z1''.
    \item Predicate names: \code{coll}, \code{cong}, \code{coll}, \code{cyclic}, \code{eqangle}, \code{eqratio}, \code{acompute}, \code{rcompute}, \code{aconst}, \code{rconst}, \code{distmeq}, \code{distseq}, \code{angeq}, \code{overlap}, \code{noverlap}, \code{sameclock}, \code{lessthan}.
    \item Number and fractions: 1, 2, 3, \dots, $-$, $/$.
    \item Predicate reference tokens: \code{(000)}, \code{(001)}, \code{(002)}, \dots \code{(999)}.
    \item Reserved tokens: \code{\{Analysis\}}, \code{\{Numerical\}}, \code{\{FromGoal\}}, \code{\{Proof\}},  \code{x00}, \code{:}, \code{;},  \code{.}.
\end{enumerate}
Somewhat surprisingly, we find that AlphaGeometry's performance on the 2000-2024 IMO geometry problems stays the same with different tokenizers, which suggests that modern LLM tokenizers might be flexible enough to perform mathematical manipulations.

\paragraph{domain-specific language.}

Alongside tokenizers, it is interesting to study the role of the domain-specific language in the LM's ability to solve math problems. It is natural to assume that using domain-specific languages simplifies mathematical manipulations and prevents obvious mistakes, which might occur from using a less strict language. To investigate this, we translate all AlphaGeometry2 data from the AlphaGeometry language into natural language and train a new model. Then we compare its performance against the model of the same size trained on the original AlphaGeometry data. Somewhat surprising again, we get the same results on 2000-2024 IMO geometry problems, which opens a path for fine-tuning large language models pre-trained in natural language on math data. Below, we demonstrate an example of translating AlphaGeometry into natural language.

AlphaGeometry language:
\code{
d e f g : coll a d g (000) coll f a b (001) coll d b c (002) coll e c a (003) cong d b d c (004) cong f a f b (005)
}

Natural language:
\code{
Construct points d e f g such that a d g are collinear (000), f a b are collinear (001), d b c are collinear (002), e c a are collinear (003), |d b| = |d c| (004), |f a| = |f b| (005)
}

\paragraph{Fine-tuning of language models pre-trained on math data.}

Having shown that the custom tokenizer and the domain-specific language does not play a critical role for AlphaGeometry, we leverage language models pre-trained on various math data. We start with a Gemini model with 3.3B parameters trained on public math datasets (see Section 7 in \cite{team2024gemini}), and fine-tune it in an unsupervised manner on the AlphaGeometry data. On our IMO-AG-50 evaluation set, the fine-tuned model performs on par with smaller models and the 3.3B model trained from scratch\footnote{We also train even larger models in a supervised manner and achieve the same results}. On the other hand, we find that even though all these models are trained on the same AG data, they do produce slightly different auxiliary points proposals and do help each other via the knowledge sharing mechanism described in Section~\ref{sec:stronger-search-algorithm}, thus forming an ensemble-like system (see Figure~\ref{fig:search-algo}).

\begin{figure*}[ht!]
    \centering
    \includegraphics[width=\textwidth]{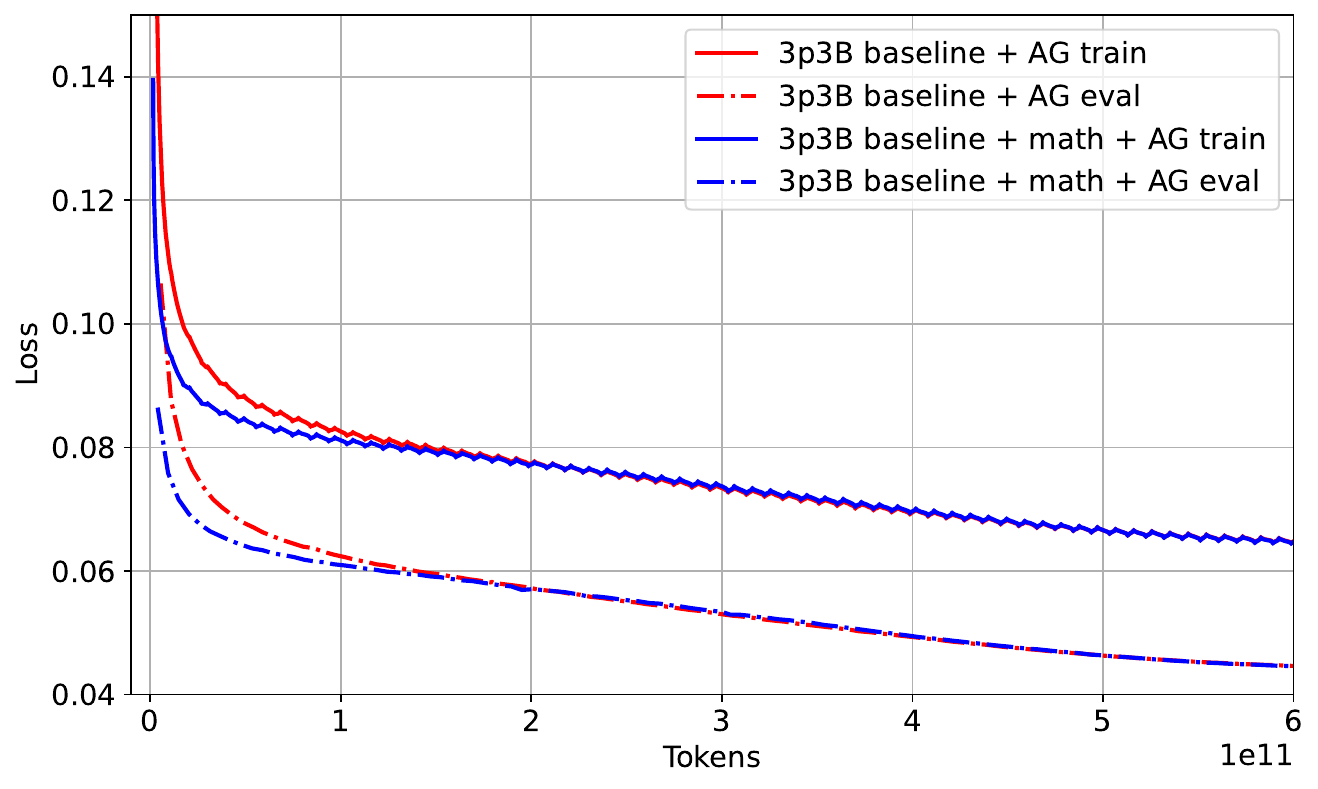}
    \caption{Learning curves for two 3B models: one is trained from scratch and another one pre-trained on math data and then fine-tuned on the AG data. The model pre-trained on math has initially lower loss but both converge to the same point after training for 200B tokens.}
    \label{fig:learning-curves-3b}
\end{figure*}

% ====================================================
% ====================================================
% ====================================================

\section{Multi-modal}
\label{app:multimodal}

Until now, we have talked about AG2 as a system that couples a language model together with a symbolic engine. However, since our language model is based on Gemini 1.5, which is multi-modal by design (see \cite{team2024gemini}), it is natural to enhance the AG system through multi-modal reasoning. For this, we train a new family of models that, alongside the problem text, take the corresponding diagram image as the input.
% For training and during test time, diagrams are built as described in Appendix~\ref{app:diagram-generation}.

Despite promising results during the training, we do not observe any improvements in the solve rate on the downstream IMO problems when using this model alone. However, just like in the case of fine-tuning pre-trained models (see Section~\ref{app:fine-tuning}), we find that the multimodal model produces slightly different auxiliary point proposals. Combined with other models via the knowledge sharing mechanism (see Section~\ref{sec:stronger-search-algorithm}) this boosts the overall performance. We hypothesize that adding the image on its own might not help that much due to very complicated diagrams, which become very crowded for the IMO problems. The image tokenization process might also play a negative role, as it splits the diagram into independent sequential patches, which leads to the loss of some spatial information. Recall also that some details about the diagram are already provided through the text, as mentioned in Section~\ref{sec:analysis}, and our symbolic engine, DDAR, does have access to topological aspects of the diagram, e.g., through inspecting \code{sameclock} predicates. Furthermore, \cite{chae2024decomposingvlm} shows that vision-language models have poor atomic visual skills, which suggests that adding visual elements might not aid the geometry problem-solving process. Finally, note that the core of geometry problem-solving lies in algebraic reasoning, as previously demonstrated in \cite{Trinh2024}, rather than geometric reasoning. Many human IMO contestants can reliably solve geometry problems (including very hard problems such as IMO 2011 P6) using computational methods such as complex numbers, barycentric coordinates and trigonometry bashing, which means that visual information and diagrams are not critical to solving geometry problems.

% ====================================================
% ====================================================
% ====================================================

\section{Featured AlphaGeometry2 solutions}
\label{app:ag2-examples}

\begin{figure*}[ht!]
    \centering
    \includegraphics[width=\textwidth]{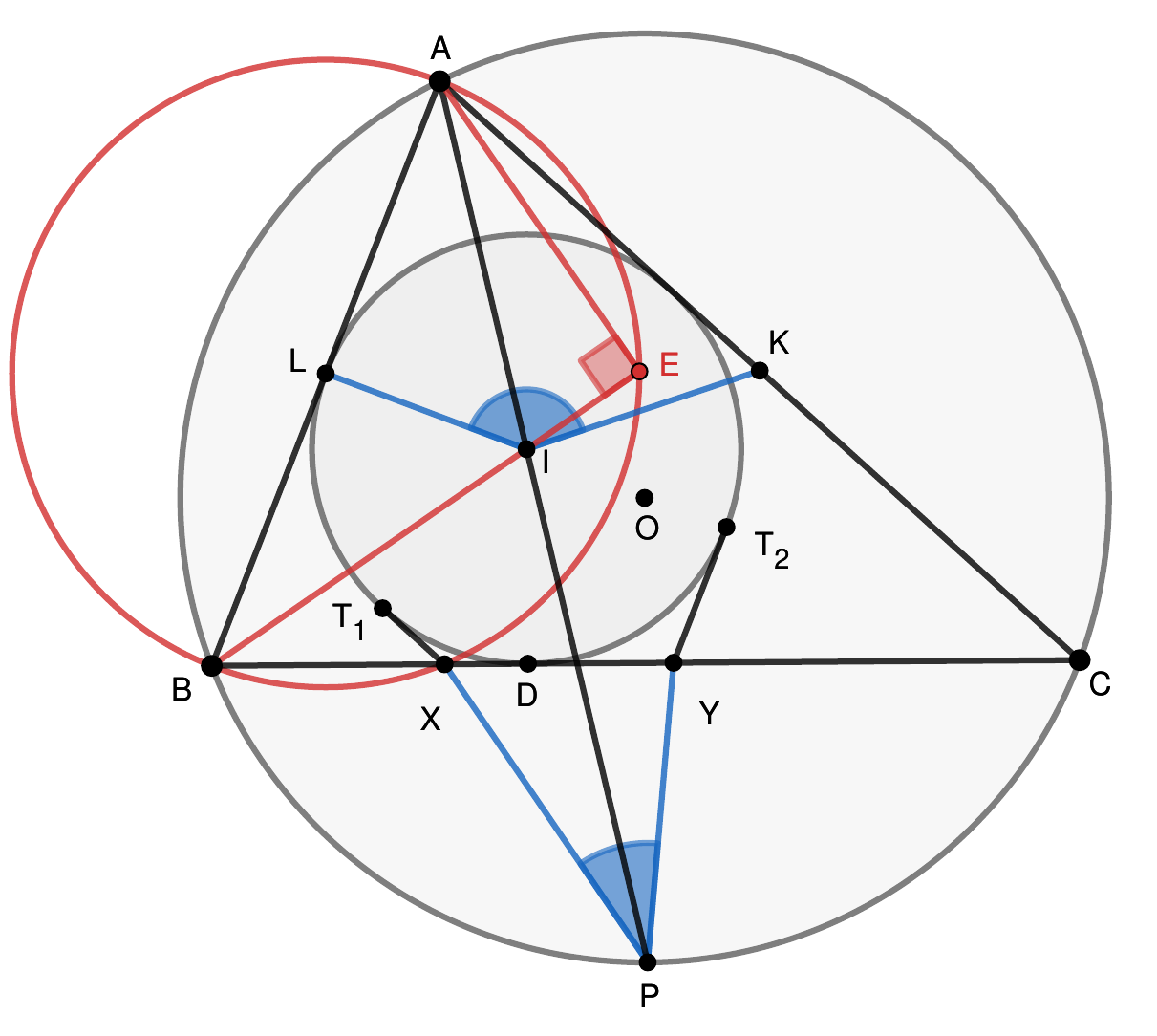}
    \caption{IMO 2024 P4 diagram with AlphaGeometry auxiliary construction, point $E$.}
    \label{fig:imo-2024-p4}
\end{figure*}

Out of the solved IMO problems (see Figure~\ref{fig:ag-solved}), our geometry experts consider many AlphaGeometry solutions to exhibit superhuman creativity. One such example is the IMO 2024 P4 problem (see Figure~\ref{fig:imo-2024-p4}).

\begin{tcolorbox}[
colback=green!5!white,
		colframe=black,
		arc=4pt,
		boxsep=0.3pt,
	]%
	\textbf{IMO 2024 P4:} Let triangle $ABC$ with incenter $I$ satisfying $AB < AC < BC$. Let $X$ be a point on line $BC$, different from $C$, such that the line through $X$ and parallel to $AC$ is tangent to the incircle. Similarly, let $Y$ be a point on line $BC$, different from $B$, such that the line through $Y$ and parallel to $AB$ is tangent to the incircle. Line $AI$ intersects the circumcircle of triangle $ABC$ again at $P$. Let $K$ and $L$ be the midpoints of $AC$ and $AB$, respectively. Prove that $\angle KIL + \angle YPX = 180^{\circ}$.
\end{tcolorbox}

The problem asks about the relationship between $\angle KIL$ and $\angle XPY$. The former is the angle formed by a midpoint and the incenter, which usually does not go well together and cannot be computed by the angles of the main triangle $ABC$. Typically, a human contestant would rely on trigonometry, complex numbers or other computational methods to find the solution. For AlphaGeometry, its DDAR system only relies on simple angle chasing and ratio chasing, so this necessitates the need for some auxiliary point constructions. To this end, AlphaGeometry constructs $E$ as a point on the line BI such that $\angle AEB = 90^{\circ}$, which elegantly ties these seemingly unrelated geometric elements together by creating pairs of similar triangles $ABE$ and $YBI$, $ALE$ and $IPC$. These pairs of similar triangles create new pairs of equal angles and equal side length ratios. 
That being said, the point $E$ gives purposes to the midpoint $L$ of $AB$. To complete the proof, we need to prove $\angle AIK = \angle BYP $ and $\angle AIL = \angle CPX$. To this end, we need to prove that triangle $AKI$ is similar to triangle $BPY$ and triangle $ALI$ is similar to triangle $CPX$, which is done by side length ratio chasing, which is obtained from the pairs of similar triangles above. A full solution is published at \url{ https://storage.googleapis.com/deepmind-media/DeepMind.com/Blog/imo-2024-solutions/P4/index.html}. This solution was obtained within 30 seconds at IMO 2024 and was given the full seven points by Joseph Myers, a two-time IMO gold medalist and Chair of the IMO 2024 Problem Selection Committee.

\begin{figure}[ht!]
\centering
\includegraphics[width=\textwidth]{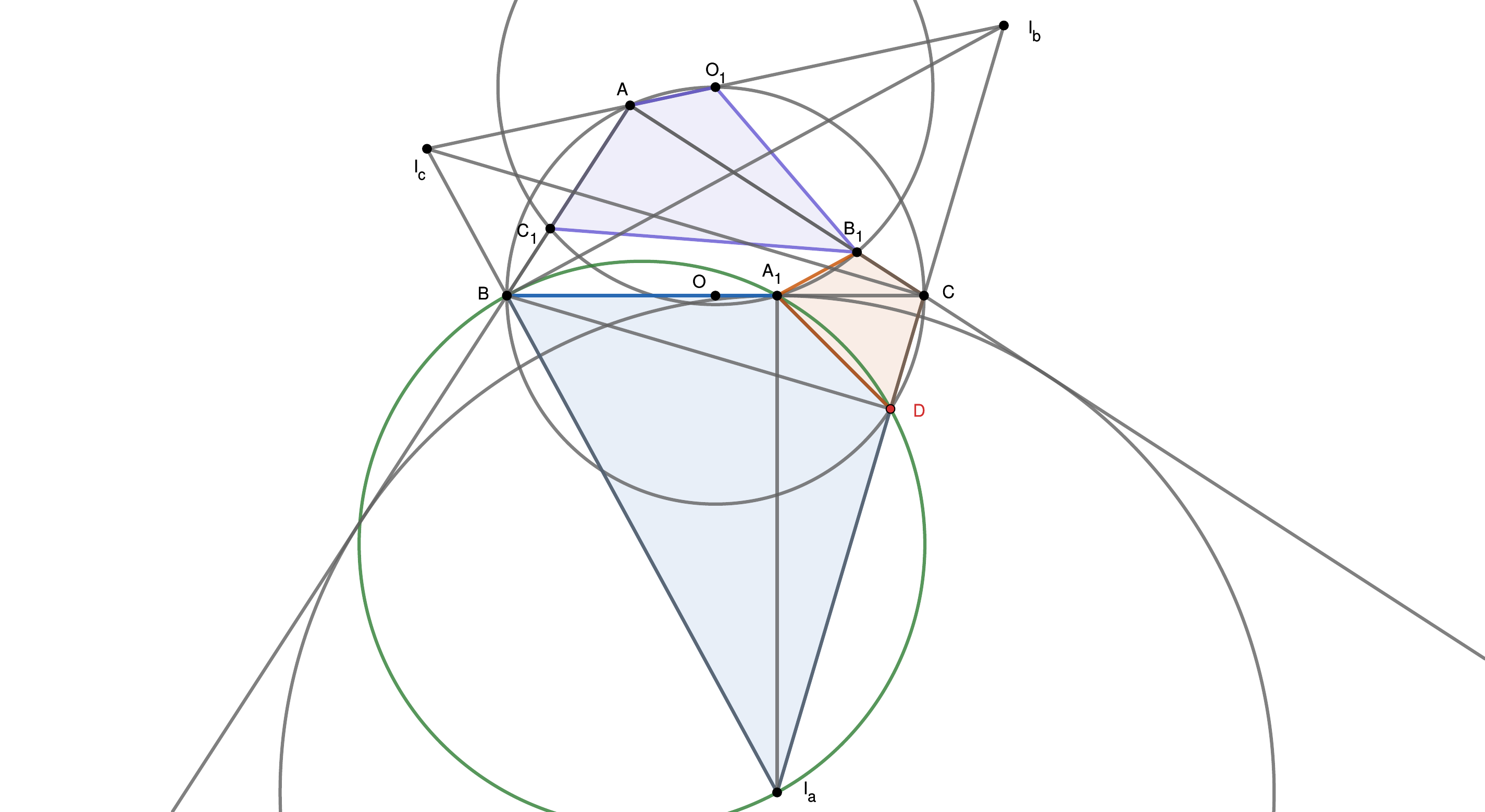}
\caption{IMO 2013 P3 diagram with AlphaGeometry auxiliary construction, point $D$. It allows proving $BA_1DI_a$ is cyclic, which is the key to solve this problem.}
\end{figure}

Along with IMO 2024 P4, AlphaGeometry can solve many challenging problems with only 1 extra auxiliary point, some of which involve rather unconventional constructions. One such problem is IMO 2013 P3.

\begin{tcolorbox}[
colback=green!5!white,
		colframe=black,
		arc=4pt,
		boxsep=0.3pt,
	]%
	\textbf{IMO 2013 P3:} Let the excircle of triangle $ABC$ opposite the vertex $A$ be tangent to the side $BC$ at the point $A_1$. Define the points $B_1$ on $CA$ and $C_1$ on $AB$ analogously, using the excircles opposite $B$ and $C$, respectively.
Suppose that the circumcenter of triangle $A_1B_1C_1$ lies on the circumcircle of triangle $ABC$. Prove that triangle $ABC$ is right-angled.
\end{tcolorbox}

In this problem, AlphaGeometry simply takes the midpoint $D$ of arc $\hat{ABC}$ containing $B$ as the extra point, which is a highly unconventional construction as it is non-symmetric. Yet, it allows AlphaGeometry to uncover the fact that $B, A_1, D, I_a$ are concyclic points, which is a key result that is only true if and only if $AB \perp AC$. To prove this fact, AlphaGeometry exploits the fact that $O_1$ and $D$ give rise to the similar triangle pairs $\triangle O_1C_1B_1 \sim \triangle O_1BC$ and $\triangle DA_1B_1 \sim \triangle DBA$ and then uses these results to facilitate angle chasing, which gives $\angle DA_1I_a = \angle DBI_a$ and the fact that $B, A_1, D, I_a$ are concyclic points follows.

\begin{figure}[ht!]
\centering
\includegraphics[width=\textwidth]{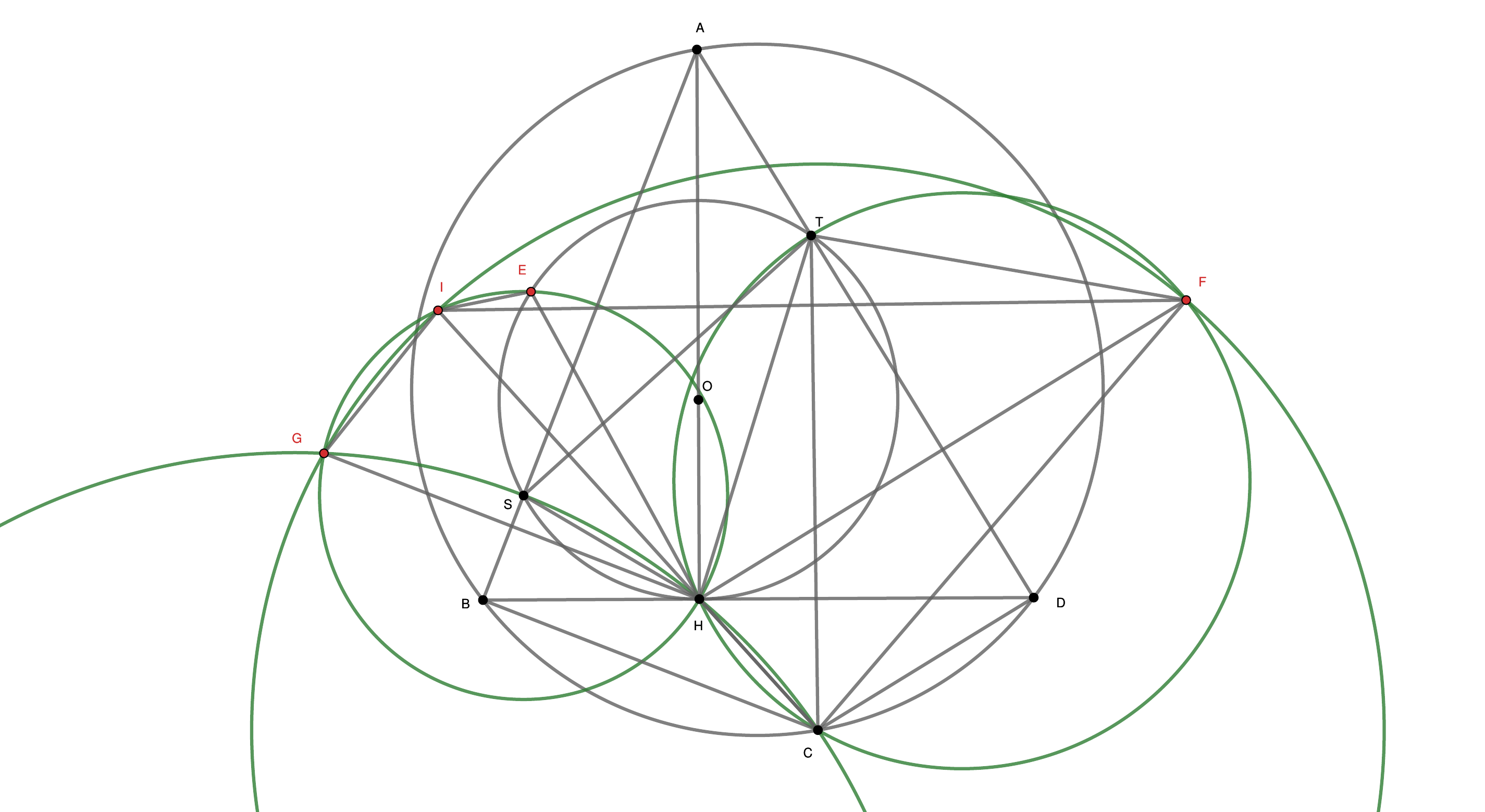}
\caption{IMO 2014 P3 diagram with AlphaGeometry auxiliary constructions.}
\end{figure}

Another example is IMO 2014 P3, one of the hard geometry problems given at the IMO.

\begin{tcolorbox}[
colback=green!5!white,
		colframe=black,
		arc=4pt,
		boxsep=0.3pt,
	]%
	\textbf{IMO 2014 P3:} Convex quadrilateral $ABCD$ has $\angle ABC = \angle CDA = 90^{\circ}$. Point $H$ is the foot of the perpendicular from $A$ to $BD$. Points $S$ and $T$ lie on sides $AB$ and $AD$, respectively, such that $H$ lies inside triangle $SCT$ and $\angle CHS - \angle CSB = 90^{\circ}, \angle THC - \angle DTC = 90^{\circ}.$ Prove that line $BD$ is tangent to the circumcircle of triangle $TSH$.
\end{tcolorbox}

To our surprise, AlphaGeometry manages to prove a more generalized result $OH \perp BD$, which implies that the circumcircle of $\triangle HST$ touches $BD$ when combining with the condition $H \in BD$ in the original problem. To do this, AlphaGeometry constructs points $E,F,G,I$ as reflections of $S$ w.r.t $OH$, $H$ w.r.t $AT$, $H$ w.r.t $AS$, $H$ w.r.t $ST$ respectively. 
Since the given conditions $\angle CHS - \angle CSB = 90^{\circ}, \angle THC - \angle DTC = 90^{\circ}$ imply that the circumcenters of $\triangle CHS, \triangle CHT$ lie on $AB, AD$ respectively, the constructions of $F$ and $G$ create cyclic quadrilaterals $CHSG, CHTF$, which will facilitate angle chasing. Moreover, the constructions of $E$ and $I$ create the cyclic quadrilateral $HGIE$ with center $S$, and the points $C,T$ now become the circumcenter of $\triangle FHI$ and $\triangle FGI$ respectively. Combining these facts altogether, AlphaGeometry obtains an extraordinary angle chasing proof, in contrast to the common approaches using ratio chasing (possibly combined with the knowledge of Apollonius circles), trigonometry or inversion by most human contestants. This shows that AlphaGeometry is capable of solving hard problems with only a simple deduction engine.

\begin{figure}[ht!]
\centering
\includegraphics[width=\textwidth]{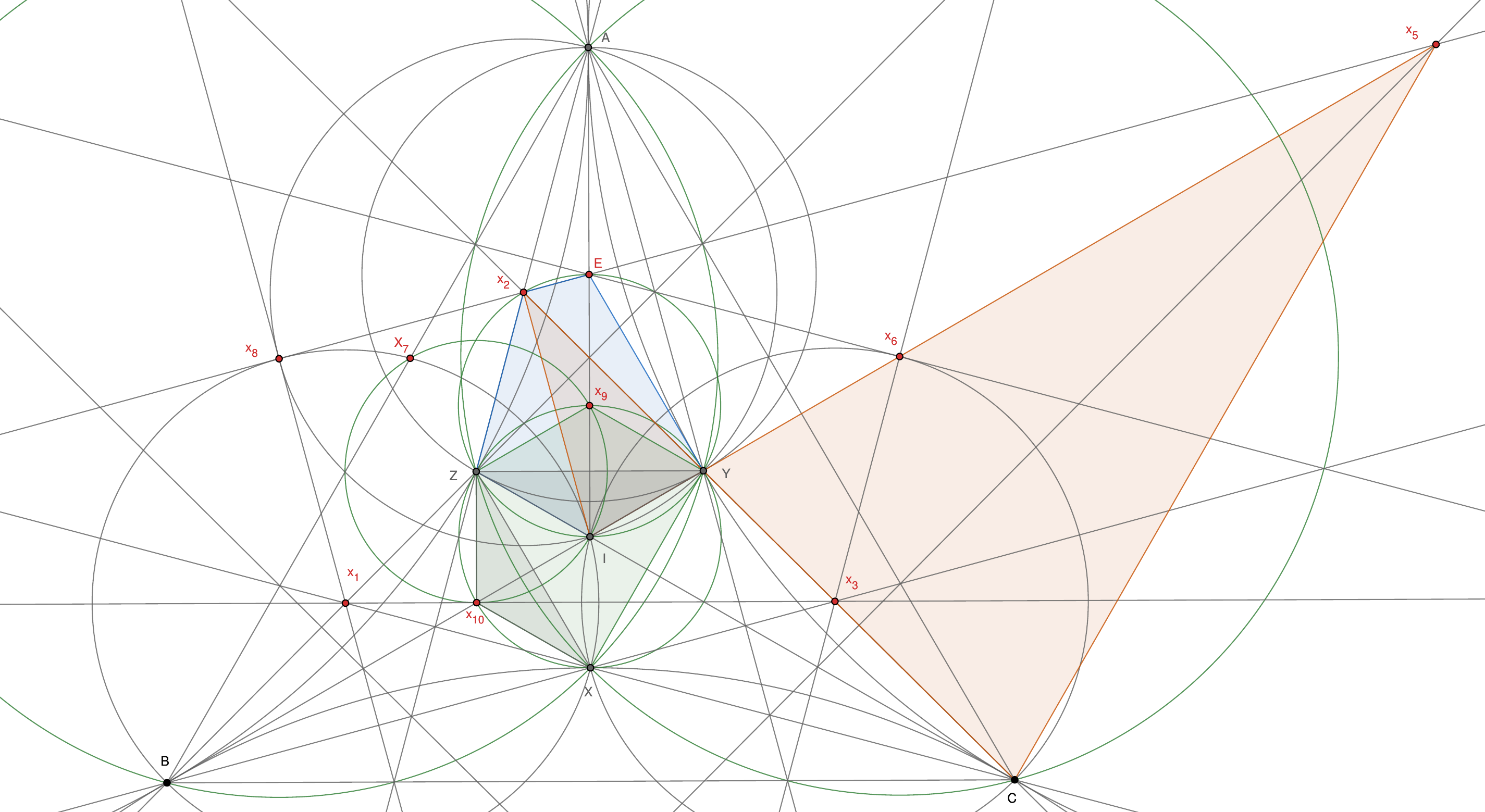}
\caption{IMOSL 2009 G7 diagram with AlphaGeometry auxiliary constructions (colored red), key cyclic properties (colored polygons) and key similar triangle pairs (colored triangle pairs).}
\end{figure}

Our final example is the G7 problem from the IMO 2009 Shortlist.

\begin{tcolorbox}[
colback=green!5!white,
		colframe=black,
		arc=4pt,
		boxsep=0.3pt,
	]%
	\textbf{IMOSL 2009 G7:} Let $ABC$ be a triangle with incenter $I$ and let $X$, $Y$ and $Z$ be the incenters of the triangles $BIC$, $CIA$ and $AIB$, respectively. Let the triangle $XYZ$ be equilateral. Prove that $ABC$ is equilateral too.
\end{tcolorbox}

To the best of our knowledge, this problem previously had only computational solutions, e.g., by using complex numbers, trigonometric computations or proof by contradiction via an inequality argument. Since AlphaGeometry does not have access to these computational and reasoning tools, as well as advanced Euclidean geometry knowledge, we originally expected that this problem cannot be solved by AlphaGeometry. Nevertheless, AlphaGeometry was able to produce an elegant solution with only angle and ratio chasing by constructing key auxiliary constructions. First, AlphaGeometry shows that $X$ and $Z$ are reflections of each other w.r.t. $BI$, and by symmetry it follows that $I$ is the circumcenter of $\triangle XYZ$. From this we can show that $AB = AC$, and by symmetry, we have $\triangle ABC$ is an equilateral triangle. However, the main challenge with this problem is to use the condition $\triangle XYZ$ being an equilateral triangle, i.e., $XY = YZ$ and its cyclic variants. To this end, AlphaGeometry constructs a series of circumcenters of key triangles:
\begin{enumerate}
    \item $D$ as the circumcenter of $\triangle BXC$.
    \item $E$ as the circumcenter of $\triangle AYZ$.
    \item $X_1$ as the circumcenter of $\triangle BIX$.
    \item $X_2$ as the circumcenter of $\triangle AIY$.
    \item $X_3$ as the circumcenter of $\triangle CIX$.
    \item $X_4$ as the circumcenter of $\triangle ABZ$.
    \item $X_5$ as the circumcenter of $\triangle ACY$.
    \item $X_6$ as the circumcenter of $\triangle AXZ$ (which we will later show that $A,C,X,Z$ are concyclic).
    \item $X_7$ as the reflection of $I$ w.r.t $BZ$
    \item $X_8$ as the circumcenter of $\triangle AXY$ (which we will later show that $A,B,X,Y$ are concyclic).
    \item $X_9, X_{10}$ are points such that $\triangle IZX_9, \triangle IZX_{10}$ are equilateral triangles.
    \item $X_{11}$ as the reflection of $Z$ w.r.t $BI$ (which is shown to be equivalent to $X$ using the point substitution technique described in Section \ref{subsec:double-points}).
\end{enumerate}
At first, these constructions seem very counter-intuitive since most humans would not construct these points. Given the nature of the points $X,Y,Z$, there are not many geometric properties related to these points and this particular configuration as a whole, which makes this problem very hard for humans to come up with a synthetic solution. Nevertheless, these circumcenter constructions help facilitate pairs of equal/similar triangles, which allow AlphaGeometry to exploit the fact that $\triangle XYZ$ is an equilateral triangle and solve the problem.

All these examples demonstrate that AlphaGeometry is very efficient in constructing auxiliary points and can offer rather elegant solutions to hard problems without using highly complex Euclidean geometry knowledge and machinery. As such, it leads to creative and efficient solutions that humans normally would not come up with.

% ====================================================
% ====================================================
% ====================================================

\section{Additional evaluation on the hardest IMO shortlist problems}
\label{app:imosl-results}

\begin{figure*}[ht!]
    \centering
    \includegraphics[width=1.0\textwidth]{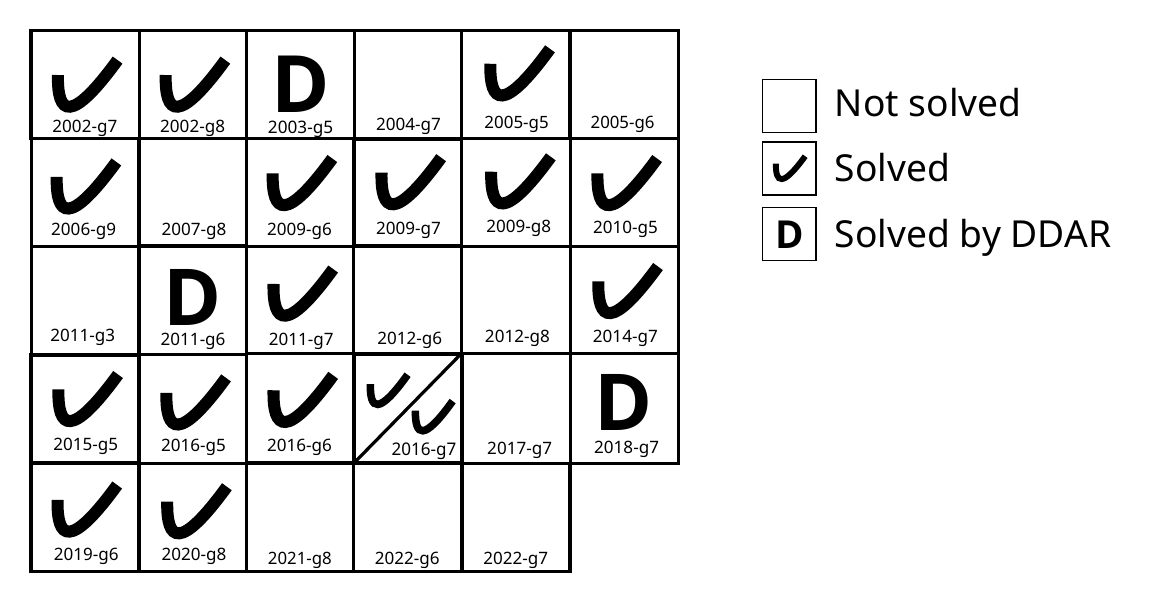}
    \caption{AlphaGeometry2 results on the hardest IMO shortlist problems.}
    \label{fig:ag-solved-imosl}
\end{figure*}

To further investigate the robustness of AG2, we perform additional evaluations on problems from the IMO shortlist that were nominated by experts but have never been selected for the IMO. Since the shortlist problems are sorted by difficulty, we select 29 problems from the end of each year's IMO shortlist that did not appear at the IMO from 2002 to 2022. These problems are selected such that they can be formalized in the AG2 language. After formalization we get 30 problems and call it IMOSL-AG-30. As demonstrated on Figure~\ref{fig:ag-solved-imosl}, the full AG2 system (see Figure~\ref{fig:search-algo}) solves 20 out of 30 problems. This shows that even though AG2 is a very capable system that can solve a wide range of Olympiad geometry problems, there is still a room for future improvements.

% ====================================================
% ====================================================
% ====================================================

\section{Towards generating full proofs with a language model}
\label{app:full-proof}

As discussed in the rest of the paper, our inference setup utilizes a language model only to produce auxiliary points followed by a symbolic engine run. On the other hand, the model is trained on whole proofs, so it is natural to ask how good the model is at generating full proofs without using the symbolic engine. Given that with greedy decoding, our model + DDAR can only solve 2 IMO problems, it is not surprising that without any further tuning, the model cannot generate complete full proofs. But can it generate partial solutions? To investigate this question we build tools to verify validity of each deduction proof step. Namely, we isolate the predicates in the premise of the proof step, and add them to a new DDAR engine, then run a deduction closure with respect to only the deduction rule being used in that step. If the new DDAR manages to prove the conclusion of the step, and the numerical check of the conclusion in the diagram passed, the step is considered verified.

Our step verification recognizes the following errors:
\begin{itemize}
    \item Wrong grammar: The step has wrong grammar.
    \item Theorem name error: The step refers to a name of a theorem (deduction rule) that does not exist.
    \item Step reference error: The step refers to a previous step that does not exist.
    \item Point not found error: The step refers to a point that does not exist, or a point with an invalid construction.
    \item Numerical error: DDAR fails due to numerical instability.
    \item Unverified: The premises of the step do not imply the conclusion of the step under the deduction rule that it uses.
    \item Invalid auxiliary point: The auxiliary point is invalid because the step is wrong in grammar, or it is geometrically invalid (e.g. intersection of two parallel lines, etc.)
    \item Verified: all checks passed and no error from above found.
\end{itemize}

\begin{figure*}[ht!]
    \centering
    \includegraphics[width=\textwidth]{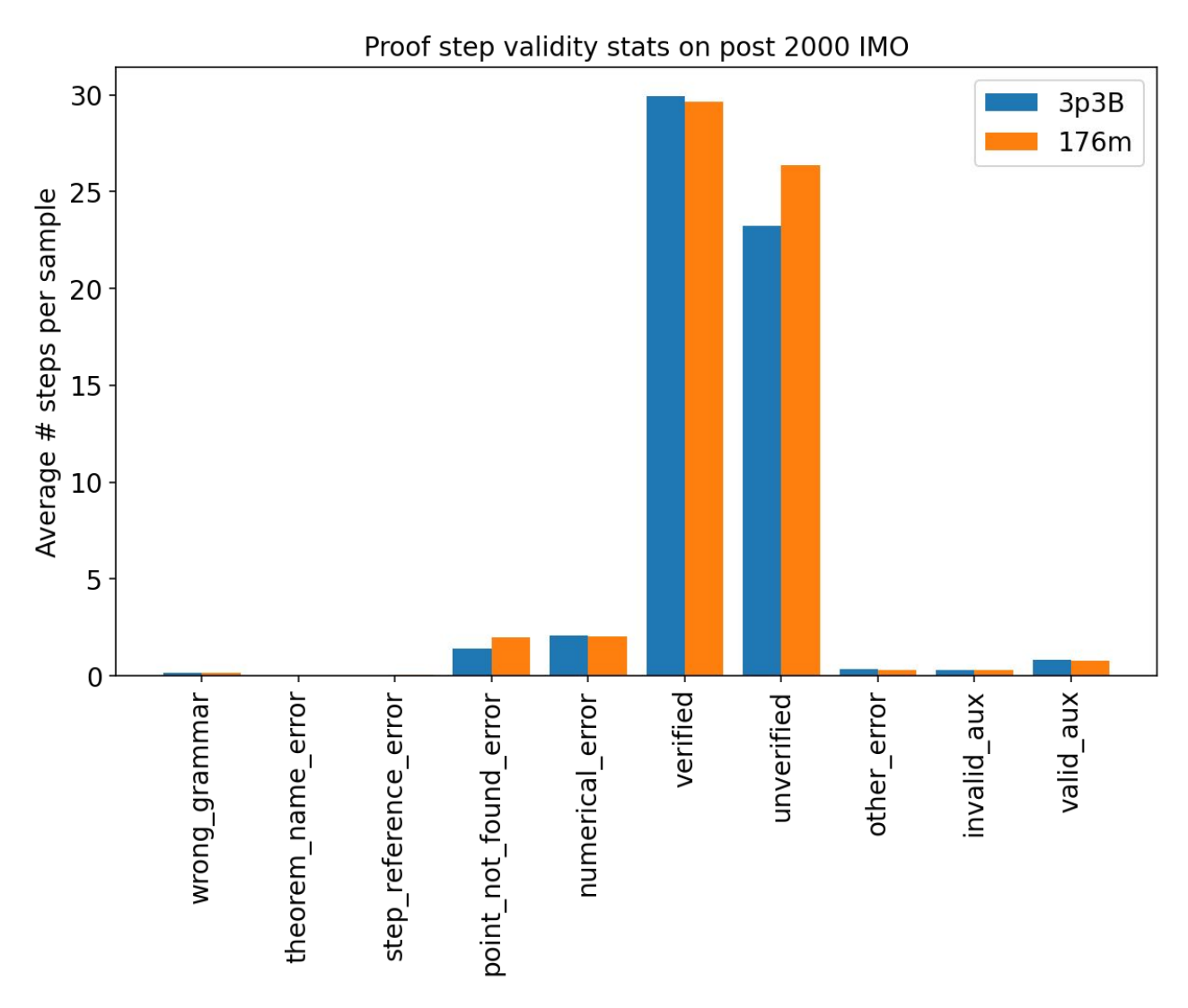}
    \caption{Proof steps validity statistics. Models almost do not make any syntax errors. Small and larger models perform similarly.}
    \label{fig:proof-steps-validaty}
\end{figure*}

For evaluation, we query our language models with the 2000-2024 IMO problems and 32 samples at temperature 1.0. The models are queried without any end-of-sentence tokens such that they generate full proofs. Then we compute how many valid proof steps the models produce on average across samples and all problems. It turns out our models do not make many syntax mistakes (see Figure~\ref{fig:proof-steps-validaty}), the majority of generated steps are either valid (either fully verified or correct but unverified). One surprising find is that both small and larger models perform similarly. These results support the idea that large language models can be self-sufficient without depending on external tools, but until inference speed is improved and hallucinations are completely resolved, the tools will stay essential for math applications.

% ====================================================
% ====================================================
% ====================================================

\section{Automated problem formalization and diagram generation}
\label{app:diagram-generation}

In this section we describe our progress in building a fully-automated AG2 system taking input in natural language and outputting the full proof along with automatically constructed diagrams. This system was used at IMO 2025 to solve the geometry problem (P2) from natural language in 20 seconds.

\paragraph{Automated formalization.} A major weakness of AlphaGeometry, and other similar neuro-symbolic systems, is the need to manually transform input problems from natural language into a domain-specific language. For example, a simple geometry problem in natural language, {\it ``Given a triangle $ABC$ with two equal sides $AB=AC$, prove that angles $B$ and $C$ are equal''}, becomes \code{triangle a b c; a b = a c ? eqangle b a b c c b c a} in the AlphaGeometry domain language.

Automating this process, called \textit{formalization}, is an active area of research (see for example, \cite{szegedy2020promising, wu2022autoformalizationlargelanguagemodels, jiang2023multilingual, poiroux2024improving, murphykaiyu2024autoformalizinggeometry}). It is a significantly more complicated problem compared to a translation between human languages. While translation aims to preserve meaning, formalization frequently requires re-formulating the original problem into an alternative form, and sometimes disambiguating the nuances in the original problem statement. Automated formalization (auto-formalization), therefore, demands significant background knowledge and problem-solving skills on its own. Given that recently foundation models started demonstrating such capabilities, we use one such model, Gemini \cite{team2024gemini}, to automate the problem formalization for AlphaGeometry. We start by manually translating several dozens of geometry problems into the AG language. Then we use these examples to write a few-shot prompt asking Gemini to translate a given geometry problem from natural language into the AG language. We query Gemini five times with this prompt, followed by another Gemini call asking to combine these results into one final answer. With this approach, we are able to formalize 33 out of 44 formalizable IMO 2000-2024 geometry problems. For easier geometry problems, it is very consistent and makes almost no mistakes. 
% In contrast to previous works, \cite{murphykaiyu2024autoformalizinggeometry} is the closest to our work where it provides formalizations of Euclidean Geometry problems in Lean, whereas our auto-formalizer translates to our domain language.

\paragraph{Automated diagram generation.} Another manual part of our pipeline was \textit{diagram generation}. In AG1, each point is defined by at most two basic predicates recalled in Table~\ref{tab:ag1-predicates}, the problem is therefore defined constructively and diagrams can be generated automatically. In AG2, we allow one or multiple points to be defined simultaneously by an arbitrary number of predicates, allowing us to also cover non-constructive problems. Consider a non-constructive problem statement, {\it``Let $ABC$ be a triangle with incenter $I$, such that $IA = 2IB$ ...''}, here point $I$ is not only defined as an incenter, i.e. the intersection of two internal bisectors, but also defined by a third predicate $IA=2IB$ and there is no general strategy to construct such four points.
Since AG2 covers non-constructive problems, diagram construction becomes a non-trivial part of the pipeline and generally requires human intervention. Similar to \cite{krueger2021automatically}, below we propose a new algorithm to automatically generate diagrams given non-constructive problem specifications.

We start by initializing the points. We have three initialization methods, which we alternate in different attempts:
\begin{enumerate}
    \item Random -- the points are chosen randomly using the normal distribution.
    \item Construction in order.
    \item Construction in a heuristically chosen order.
\end{enumerate}

In order to construct a new point $X$, we look at each predicate $p(X, A_1, \ldots, A_k)$ involving $X$ point, and already constructed points $A_1, \ldots, A_k$, and find a circle or line that X must be on.
For example, if we know $|A_1A_2| = |A_3X|$, we have already constructed, we know $X$ should be on a circle with center $A_3$, and radius $|A_1A_2|$. If we cannot obtain such a circle, or a line, we skip that predicate. If there is not such set, $X$ is sampled randomly with a normal distribution. If there is exactly one such set, $X$ is sampled from it. If there are more such sets, $X$ is sampled to be any intersection of any two of them. We sample 10 examples for $X$, and choose one which minimizes the loss described below (with added noise to improve exploration).

After point initialization, we optimize their coordinates.
Let $\bar x \in \mathbb{R}^{2n}$ be a vector representing all coordinates of all points. We encode every exact constraint $c$ in the diagram, including the goal, as $f_c(\bar x) = 0$ with a nonlinear function $f_c$, and each topological constraint as $g_c(\bar x) < 0$, or $h_c(\bar x) \neq 0$. We numerically search for a suitable $\bar x$ in two steps. The overall loss for the Adam optimizer consists of
\begin{enumerate}
    \item $\displaystyle\sum_{c\in C} f_c(\bar x)^2$ where $C$ is the set of all the exact constraints,
    \item $\displaystyle\sum_{c\in C} {\rm softplus}(g_c(\bar x))$ where $C$ is the set of all the topological constraints requiring $g_c(\bar x) < 0$,
    \item $\displaystyle\sum_{c\in C} {\rm softplus}(\min(h_c(\bar x), -h_c(\bar x))$ where $C$ is the set of all the topological constraints requiring $h_c(\bar x) = 0$,
    \item non-degeneracy loss of the form of a fraction of $L_2$ norm of all the point coordinates,
    \item non-degeneracy loss of the form of $1/{(|AB|^2 + \epsilon)}$ for all pairs of distinct points $A, B$.
\end{enumerate}
We run the ADAM gradient descent optimization for a fixed number of steps simultaneously for 10 initial setups. Afterwards, we filter only the diagrams where the loss got below a given threshold, and all the topological constraints are satisfied. Finally, we switch from a gradient descent optimization to the Gauss-Newton-Levenberg method to look for a numerical solution of a combined under- and over-determined system of nonlinear equations.

This three-stage optimization method is similar to the methodology introduced in \cite{krueger2021automatically}. The final stage addresses the practical limitations encountered when tuning the gradient descent optimization in the original method, where achieving a consistently satisfactory error margin proved challenging.

We benchmark this method on 44 IMO problems formalized in AG language (see Figure~\ref{fig:ag-solved}) and are able to find diagrams for 43. We run the three-stage convergence procedure in a loop which restarts and generates another random initial configuration after a failure. This way, 43 / 44 problems got their diagram sequentially generated within 1 hour.

% ====================================================
% ====================================================
% ====================================================

\section{Inequality rules}
\label{app:inequality-rules}

This section details a set of inequality rules. Although potentially useful for many inequality-related questions from Euclidean geometry, these rules were not included in the AlphaGeometry2 system.
\subsection{Definitions}
\begin{itemize}
    \item $P(A_1A_2\ldots A_n)$ is a polygon formed by $A_1,A_2,\ldots, A_n$.
    \item $\omega(ABC)=1$ if $ABC$ is clockwise, $-1$ otherwise.
    \item $\eta(ABC)=1$ if $ABC$ collinear and $B$ is between $A$ and $C$, $-1$ if $A$ and $C$ are on the same side with respect to $B$.
    \item $\angle ABC$ the angle formed by the segments $AB$ and $BC$ whose measure is less than $\pi$.
    \item $\angle_0(ABC)$: traditional unoriented measure of $\angle ABC$ Ranges from $0$ to $\pi$.
    \item $\angle_1(ABC)$: angle measured from line $AB$ to $BC$ counterclockwise. Ranges from $0$ to $\pi$.
    \item $\angle_2(ABC)$: angle measured from ray $BA$ to $BC$. Positive if counterclockwise and negative if clockwise. Ranges from $-\pi$ to $\pi$.
\end{itemize}

\subsection{How to write some of the common geometric inequality statements using \texorpdfstring{$\omega$}.}
\begin{itemize}
    \item $C$ and $D$ separated by $AB$: $\omega(ABC)=\omega(ADB)$
    \item $D$ is contained in $P(ABC)$: $\omega(DBC)=\omega(ADC)=\omega(ABD) (=\omega(ABC))$
    \item $P(A_1A_2\ldots A_n)$ is convex: $\omega(A_1A_2A_3) = \omega(A_2A_3A_4)=\ldots=\omega(A_nA_1A_2)$.
    \item $p \in P(A_1A_2\ldots A_n)$ is convex: $\omega(pA_1A_2) = \omega(pA_2A_3)=\ldots =\omega(pA_nA_1)(=\omega(A_1A_2A_3) = \omega(A_2A_3A_4)=\ldots=\omega(A_nA_1A_2))$
    \item $D$ is in the (minor) sector determined by the ray $AB$ and ray $AC$: $\omega(BAD)=\omega(BAC)=\omega(DAC)$
\end{itemize}

\subsection{Trivial rules}
\begin{itemize}
    \item ncol $A~ B~ C$ $\Rightarrow$ $\omega(ABC)=\omega(BCA)$
    \item ncol $A~ B~ C$ $\Rightarrow$ $\omega(ABC) \neq \omega(ACB)$
    \item $\Rightarrow$ $\eta(ABC) = \eta(CBA)$
    \item $\omega(ABC) \neq \omega(DEF)$ $\Rightarrow$ $\omega(ABC) = \omega(DFE)$
    \item $\Rightarrow$ $\angle_0(ABC) = \omega(ABC)\angle_2(ABC)$
    \item $\Rightarrow$ $\angle_0(ABC) = |\angle_2(ABC)|$
    \item $\Rightarrow$ $\angle_1(ABC) = \angle_2(ABC) + \pi (1-\omega(ABC))/2$
\end{itemize}

\subsection{Rules for relating \texorpdfstring{$\omega$}. and \texorpdfstring{$\eta$}.}
\begin{itemize} 
    \item $\eta(ABC)=1$, ncol $X~A~B$  $\Rightarrow$ $\omega(XAB)= \omega(XBC)=\omega(XAC)$
    \item $\eta(ABC)=-1$, ncol $X~A~B$ $\Rightarrow$ $\omega(XBA)=\omega(XBC)$
    \item ncol $X~A~B$, col $A~B~C$, $\omega(XAB)=\omega(XBC)$ $\Rightarrow$ $\eta(ABC)=1$
    \item ncol $X~A~B$, col $A~B~C$, $\omega(XBA)=\omega(XBC)$ $\Rightarrow$ $\eta(ABC)=-1$
\end{itemize}

\subsection{Basic rules using P in polygon}
\begin{itemize}
    \item ncol $A_1~A_2~A_3$  $\Rightarrow$ $P(A_1A_2A_3)$: convex
    \item $\omega(A_1A_2A_3) = \omega(A_2A_3A_4)=\ldots=\omega(A_nA_1A_2)$ $\Rightarrow$ $P(A_1A_2\ldots A_n)$: convex
    \item $P(A_1A_2\ldots A_n)$: convex $\Rightarrow$ $\omega(A_1A_2A_3)=\omega(A_iA_jA_k)$ with any $i<j<k$ 
    \item $P(A_1A_2\ldots A_n)$: convex, $\omega(pA_1A_2) = \omega(pA_2A_3)=\ldots=\omega(pA_nA_1)$ $\Rightarrow$ $p \in P(A_1A_2\ldots A_n)$
    \item $p \in P(A_1A_2\ldots A_n)$ $\Rightarrow$ $\omega(pA_1A_2)=\omega(A_1A_2A_3)$
    \item $p \in P(A_1A_2\ldots A_n)$, $P(A_1A_2\ldots A_nA_{n+1})$: convex $\Rightarrow$ $p \in P(A_1A_2\ldots A_nA_{n+1})$
    \item $p \in P(ABC)$, $\eta(DBC) = 1$ $\Rightarrow$ $p \in P(ADC)$
    \item $p \in P(A_1A_2\ldots A_n)$, col $X~ A_n~ A_1$, col $X~ A_{n-1}~ A_{n-2}$  $\Rightarrow$ $p \in P(A_1A_2\ldots A_{n-2} X)$
    \item $P(A_1\ldots A_n)$: convex $\Rightarrow$ $P(A_1\ldots A_n-1)$: convex
    \item $P(ABCD)$: convex, col $A~X~C$, col $B~X~D$ $\Rightarrow$ $X \in P(ABCD)$
\end{itemize}
\subsection{Basic rules using ``\texorpdfstring{$\in O$}.'' and ``\texorpdfstring{$\notin O$}.'' }
 Here, $O(P,AB)$ is a circle centered at $P$ with the radius $AB$, and $O(ABC)$ is the circumcircle of $P(ABC)$.
\begin{itemize}
    \item col $A~O~B$, $OA = OB$, $\eta(AOB)=1$, $\angle_0 ACB< \pi /2$ $\Rightarrow$ $C \notin O(O,AO)$
    \item col $A~O~B$, $OA = OB$, $\eta(AOB)=1$, $\angle_0 ACB> \pi /2$ $\Rightarrow$ $C \in O(O,AO)$
    \item $\omega(ADB) = \omega(ACB)$, $\angle_0 ACB > \angle_0 ADB$ $\Rightarrow$ $D \notin O(ABC)$
    \item $\omega(ADB) = \omega(ACB)$, $\angle_0 ACB < \angle_0 ADB$ $\Rightarrow$ $D \in O(ABC)$
    \item $\omega(ADB) = -\omega(ACB)$, $\pi - \angle_0 ACB > \angle_0 ADB$ $\Rightarrow$ $D \notin O(ABC)$
    \item $\omega(ADB) = -\omega(ACB)$, $\pi - \angle_0 ACB < \angle_0 ADB$ $\Rightarrow$ $D \in O(ABC)$
    \item $D \notin O(ABC)$, $\omega(ADB)=\omega(ACB)$ $\Rightarrow$ $\angle_0 ACB > \angle_0 ADB$
    \item $D \notin O(ABC)$, $\omega(ADB)=-\omega(ACB)$ $\Rightarrow$  $\pi - \angle_0 ACB > \angle_0 ADB$
    \item $p \in P(ABC)$ $\Rightarrow$ $p \in O(ABC)$
    \item $p \in O(ABC)$, $\omega(pBC)=\omega(ABC)$ $\Rightarrow$ $\angle_0 BpC > \angle_0 BAC$
    \item $p \in O(ABC)$, $\omega(pBC)=-\omega(ABC)$ $\Rightarrow$ $\angle_0 BpC >\pi - \angle_0 BAC$
    \item $D \in O(O,OA)$ $\Rightarrow$ $OD<OA$
    \item $OD<OA$ $\Rightarrow$ $D \in O(O,OA)$
    \item $D \notin O(O,OA)$ $\Rightarrow$ $OD>OA$
    \item $OD>OA$ $\Rightarrow$ $D \notin O(O,OA)$
\end{itemize}

\subsection{Acute and obtuse angles}
\begin{itemize}
    \item $\angle ABC$: acute $\Rightarrow$ $\angle_0 ABC <  \pi /2$
    \item  $\pi >\angle_2(ABC)> \pi /2$, circ $O~ A~ B~ C$ $\Rightarrow$ $\omega(OAC)= \omega(BCA)$
    \item  $\pi >\angle_2(ABC)> \pi /2$, $H$: orthocenter of $P(ABC)$ $\Rightarrow$ $\omega(AHC)= \omega(ABC)$
    \item $G$: centroid of $P(ABC)$ $\Rightarrow$ $G \in P(ABC)$
    \item $I$: incenter of $P(ABC)$ $\Rightarrow$ $I \in P(ABC)$
    \item $P(ABC)$: acute $\Rightarrow$ $O, H \in P(ABC)$
    \item col $B~D~C$, ncol $A~B~D$, $AB=AD$, $\angle ABC$: acute $\Rightarrow$ $\eta(DBC)=-1$
    \item col $B~D~C$, ncol $A~B~D$, $AB=AD$, $\angle ABC$: obtuse $\Rightarrow$ $\eta(DBC)=1$
    \item col $B~D~C$, $\angle_0 (ABC)<\angle_0 (ADC)$ $\Rightarrow$ $\eta(DBC)=-1$
    \item midpoint $M~B~C$, col $A~X~B$, $XM \perp BC$, $AB>AC$ $\Rightarrow$ $\eta(AXB)=1$
    \item midpoint $M~B~C$, col $A~X~B$, $XM \perp BC$, $AB<AC$ $\Rightarrow$ $\eta(XAB)=1$
\end{itemize}

\subsection{Some inequalities}
\begin{itemize}
    \item $\eta(ABC)=1$ $\Rightarrow$ $AC>AB$
    \item $\eta(ABC)=1$ $\Rightarrow$ $AC=AB+BC$
    \item col $A~B~C$, $AC>AB$, $AC>BC$ $\Rightarrow$ $\eta(ABC)=1$
    \item $\omega(BAD)=\omega(BAC)=\omega(DAC)$ $\Rightarrow$ $\angle_0(BAD)<\angle_0(BAC)$, $\angle_0(DAC) < \angle_0(BAC)$
    \item $\angle_0(BAD)<\angle_0(BAC)$, $\angle_0(DAC)<\angle_0(BAC)$ $\Rightarrow$ $\omega(BAD)=\omega(BAC)=\omega(DAC)$ 
    \item {} $\Rightarrow$ $AB+BC  \geq AC$.  
    \item $AB+BC=AC$ $\Rightarrow$ col $A~B~C$, $\eta(ABC)=1$
    \item col $B~C~D$, col $A~O~D$, $OA=OB$, $OB=OC$, $P(ABC):$ acute, $AB<AC$ $\Rightarrow$ $BD>DC$
    \item col $B~C~D$, col $A~O~D$, $OA=OB$, $OB=OC$, $\angle_0 (BAC) > \pi /2$, $AB<AC$ $\Rightarrow$ $BD<DC$
    \item col $A~B~C$, $\eta(BAC)=-1$, $CA>BA$ $\Rightarrow$ $\eta(ABC)=1$
    \item $\eta(ACB)=1$, $\eta(ADB)=1$ $\Rightarrow$ $\eta(CAD) = -1$
    \item $AB \perp BC$ $\Rightarrow$ $AC>AB$, $AC>BC$
    \item $AB>AC$ $\Rightarrow$ $\angle_0 (ACB) > \angle_0 (ABC)$
    \item $\angle_0 (ACB) > \angle_0 (ABC)$ $\Rightarrow$ $AB>AC$
\end{itemize}

\subsection{General \texorpdfstring{$\omega$}. and \texorpdfstring{$\eta$}. deduction rules}
\begin{itemize}
    \item $\omega(ABC)=\omega(BCD)$, $\omega(BCD)=\omega(CDA)$ $\Rightarrow$ $\omega(DAB) = \omega(CDA)$
    \item $\angle_2(ABC)>0$ $\iff$ $\omega(ABC)=1$
    \item $\angle_2(ABC)<0$ $\iff$ $\omega(ABC)=-1$
    \item $\Rightarrow$ $\angle_0(ABC)+\angle_0(BCA)+\angle_0(CAB)=\pi$
    \item $\eta(BCX)=1$ $\Rightarrow$ $\angle_2 (ACX) = \angle_2 (ABC)+\angle_2(CAB)$
    \item circ $O~ A~ B~ C$, $\omega(BOA)=\omega(COB)$  $\Rightarrow$ $\omega(ABC) = \omega(BOA)$
    \item $\eta(CAB)=-1$, $\eta(ACB)=-1$ $\Rightarrow$ $\eta(ABC)=1$
    \item $\eta(ABC)=1$ $\Rightarrow$ $\eta(CAB)=-1$, $\eta(ACB)=-1$
    \item $\eta(ABC)=1$, $\eta(BCD)=1$ $\Rightarrow$ $\eta(ABD)=1$, $\eta(ACD)=1$
    \item col $B~X~C$, $AX \perp BC$, $\angle_0 (ABC) < \pi /2$, $\angle_0 (ACB) < \pi /2$ $\Rightarrow$ $\eta(BXC)=1$
    \item col $B~X~C$, $\angle_2(BAX)=\angle_2(XAC)$ $\Rightarrow$ $\eta(BXC)=1$
    \item midpoint $M~ A~ B$  $\Rightarrow$ $\eta(AMB)=1$
    \item $\eta(ADB)=\eta(AEC)=1$ $\Rightarrow$ $P(D E C B)$: convex
    \item $X \notin O(ABC)$, $\omega(XAB)=\omega(CAB)$, cyclic $A~B~C~D$, col $X~A~D$, $\angle_0(CBA)>\angle_0(CAX)$ $\Rightarrow$ $\eta(XDA)=1$
    \item $X \notin O(ABC)$, $\omega(XAB)=\omega(CAB)$, cyclic $A~B~C~D$, col $X~A~D$, $\angle_0(BCA)+\angle_0(XAB)<\pi$ $\Rightarrow$ $\eta(XDA)=1$
    \item $X \notin O(ABC)$, $\omega(XAB)=\omega(CAB)$, cyclic $A~B~C~D$, col $X~A~D$, $\angle_0(CBA)<\angle_0(CAX)$ $\Rightarrow$ $\eta(XAD)=1$
    \item $X \notin O(ABC)$, $\omega(XAB)=\omega(CAB)$, cyclic $A~B~C~D$, col $X~A~D$, $\angle_0(BCA)+\angle_0(XAB)>\pi$ $\Rightarrow$ $\eta(XAD)=1$
    \item $X \notin O(ABD)$, col $A~ O~ B$, col  $X~ A~ D$, $\angle_0(XAB)< \pi /2$ $\Rightarrow$ $\eta(XDA)=1$
    \item $X \notin O(ABD)$, col $A~ O~ B$, col  $X~ A~ D$, $\angle_0(XAB)> \pi /2$ $\Rightarrow$ $\eta(XAD)=1$
    \item $X \in O(ABC)$, col $A~X~B$ $\Rightarrow$ $\eta(AXB)=1$
    \item $\angle_2 XBA+ \angle_2 XBC = \omega(XBA) \pi$, 
    $\angle_2 XCA+ \angle_2 XCB = \omega(XCA) \pi$ 
    $\Rightarrow$ $X \notin O(ABC)$  ($X$ will be the excenter relative to $A$: $X==I_A$)
\end{itemize}

\end{document}